\documentclass[10pt,journal,compsoc]{IEEEtran}
 
\ifCLASSOPTIONcompsoc 
 \usepackage[nocompress]{cite}
\else 
 \usepackage{cite}
\fi
 
\ifCLASSINFOpdf
\else
\fi

\usepackage{comment}
\usepackage{amsmath} 
\usepackage{eso-pic,xspace}
\usepackage{amssymb}
\usepackage{bm} 
\usepackage{color}
\usepackage{floatrow}
\usepackage{array}
\usepackage{xcolor,colortbl}	
\usepackage{epsfig}
\usepackage{graphicx}
\usepackage{tabularx,ragged2e}
\usepackage[export]{adjustbox}
\usepackage{booktabs,calc}
\usepackage{multirow}
\usepackage{hhline}
\usepackage{float} 
\usepackage{subfig}
\usepackage[scientific-notation=true]{siunitx}
\usepackage{type1cm}
\usepackage{lettrine}
\usepackage{rotating} 
\usepackage{longtable}
\usepackage{wrapfig}
\usepackage{colortbl}
\usepackage{stmaryrd} 
\usepackage{setspace}
\usepackage{url} 
\usepackage{comment}
\usepackage{tabularx}
\usepackage{cuted}
\usepackage{capt-of}
\usepackage{xspace} 
\usepackage{hyperref}
\hypersetup{
 colorlinks=true,
 linkcolor=red,
 citecolor=green,
}
 
\definecolor{greenff}{rgb}{0.0, 0.5, 0.0} 
 
\makeatletter
\DeclareTextFontCommand{\emph}{\em} 
\DeclareRobustCommand\onedot{\futurelet\@let@token\@onedot}
\def\@onedot{\ifx\@let@token.\else.\null\fi\xspace}
\def\eg{e.g\onedot} 
\def\ie{i.e\onedot} 
 
\def\etc{etc\onedot}  
 
\def\etal{et al\onedot}

\begin{document}
 
\title{Transferring Knowledge from Text to Video: Zero-Shot Anticipation for Procedural Actions}
 
\author{Fadime Sener, 
 Rishabh Saraf, 
 and~Angela Yao 
\IEEEcompsocitemizethanks{

\IEEEcompsocthanksitem Fadime Sener is with the University of Bonn
\protect\\
E-mail: sener@cs.uni-bonn.de

\IEEEcompsocthanksitem Rishabh Saraf is with Indian Institute of Technology (ISM) Dhanbad
\protect\\
E-mail: rishabh.15je001745@am.iitism.ac.in

\IEEEcompsocthanksitem Angela Yao is with National University of Singapore, Singapore
\protect\\
E-mail: ayao@comp.nus.edu.sg
}
}

\IEEEtitleabstractindextext{
\begin{abstract}
Can we teach a robot to recognize and make predictions for activities that it has never seen before? We tackle this problem by learning models for video from text. This paper presents a hierarchical model that generalizes instructional knowledge from large-scale text corpora and transfers the knowledge to video. Given a portion of an instructional video, our model recognizes and predicts coherent and plausible actions multiple steps into the future, all in rich natural language. To demonstrate the capabilities of our model, we introduce the \emph{Tasty Videos Dataset V2}, a collection of 4022 recipes for zero-shot learning, recognition and anticipation. Extensive experiments with various evaluation metrics demonstrate the potential of our method for generalization, given limited video data for training models. 
\end{abstract}

}
 
\maketitle

\IEEEdisplaynontitleabstractindextext

\IEEEraisesectionheading{\section{Introduction}\label{sec:intro}}

\IEEEPARstart{I}magine a not-so-distant future where robot chefs service our kitchens. How can we learn and embody cooking as a general skill? Perhaps by reading all the recipes on the web? Or by watching all the cooking videos on YouTube? Learning and generalizing from a set of instructions, be it in text, image, or video form, is a highly challenging and open problem faced by those working in computer vision, natural language understanding and robotics. 

In this work, we limit our scope of training the next `robochef' to predict subsequent steps as it watches a human cook a never-before-seen dish. We frame the problem as action recognition and anticipation in a zero- or few-shot learning scenario. In addition to recognition, it will be important for the robot to anticipate what happens in the future to ensure a safe and smooth collaborative experience with the human~\cite{koppula2015anticipating,wu2016watch}. We also place importance on zero-shot or few-shot learning as it likely reflects how service robots will be introduced to the home~\cite{Chelsea04905,sunderhauf2018limits}. Models (and robots) can be pre-trained extensively, but they will likely be deployed in never-before-encountered scenarios. Successfully anticipating the next steps of a never-before-seen dish would require leveraging and generalizing from previously learned procedural knowledge. 

Instructional data, especially cooking recipes, can be found readily on the web~\cite{tang2019coin,Wikihow,zhukov2019cross}. The richest forms are multimodal \eg, images plus text or videos with narrations. Such data could be used to build automated systems to learn visual models from videos, enabling the advancement of virtual assistants or service robots learning new skills. However, learning complex multi-step procedures requires significant amounts of data. Despite the abundance of instructional data online, it is still difficult to find sufficient examples in multimodal form. Furthermore, learning the steps' visual appearance would require temporally aligned data, which is less common and or expensive to annotate. Several works~\cite{miech2019howto100m,miech2020end} have been proposed to tackle learning video representations without manual supervision. However, these methods can only provide solutions for coping with the misaligned narrations and cannot handle missing or unrelated narration. 

\begin{figure*}[htb!]
\centering 
\includegraphics[width=1\textwidth]{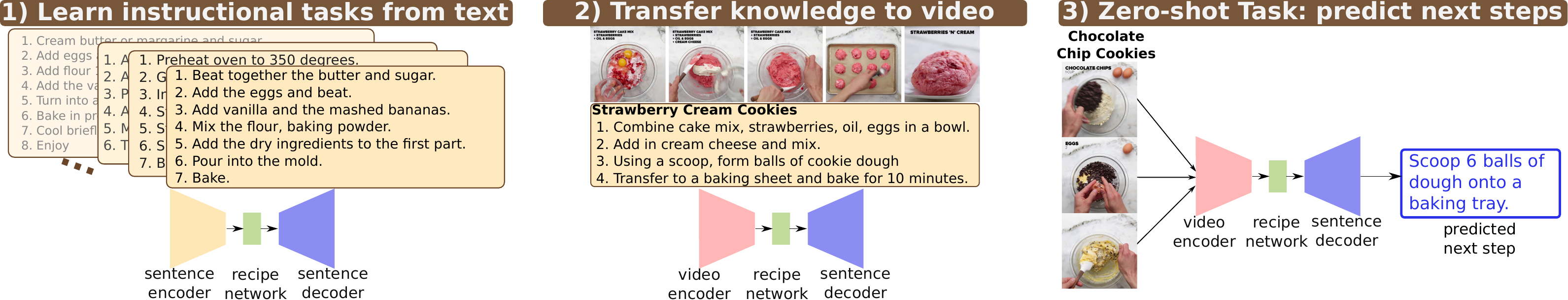} 
\captionof{figure}{
An overview of our model. We first learn procedural knowledge from large text corpora and transfer it to the visual domain to anticipate the future. Our system comprises four RNNs: a sentence encoder, a sentence decoder, a video encoder, and a recipe network.
}
\label{fig:teaser}
\end{figure*}

Our strategy is to separate the procedural learning from the visual perception problem. We learn procedural knowledge from text corpora; these are readily available and large scale, on the scale of millions~\cite{salvador2017learning,malmaud2014cooking,miech2019howto100m}. Knowledge from the text is then transferred to video so that visual perception is simplified to a grounding task done via aligned video and text (Fig.~\ref{fig:teaser}).
Early examples of transferring knowledge from one domain to another like~\cite{ngiam2011multimodal,srivastava2012multimodal} work with images, sound, and text. Ngiam~\etal~\cite{ngiam2011multimodal} represent image and sound using separate stacked autoencoders and fuse them into a multimodal representation space. Srivastava and Salakhutdinov~\cite{srivastava2012multimodal} learn a joint space for image and text for information retrieval from unimodal or multimodal queries. More recently, VideoBERT~\cite{sun2019videobert} has been proposed for the joint modeling of video and text. Unlike these works, which assume parallel data during training, our text models are trained on large text corpora alone, and the video models are trained on scarce parallel data and can be used for zero-shot queries.

More specifically, we encode text and or video in a multimodal embedding space. The context vectors, derived from either video or text, are fed into a recipe network that models the recipe's sequential structure and makes the following step predictions. Figure~\ref{fig:teaser} shows an overview. The use of text and language information to help train video models is not new. Several works employ accompanying narrations~\cite{zhukov2019cross}, recipes~\cite{malmaud2014cooking} or film scripts~\cite{zhu2015aligning} to use text as weak boundaries and several learn joint text-video models~\cite{miech2019howto100m,sun2019videobert}. Our work is similar in spirit in that we also want to leverage these auxiliary sources of information to reduce the labeling effort. However, the previous works mainly focus on using text to minimize the labeling effort for large-scale video datasets. In contrast, our work learns entire models out of text and then transfers said models across the natural language domain to the visual domain via aligned video and text representations.
To the best of our knowledge, we are the first to learn and transfer a cross-domain model. 

Our work breaks new ground in procedural activity understanding in two ways. First and foremost, we anticipate upcoming actions under a zero-shot setting, as we target making predictions for never-before-seen dishes. We achieve this by generalizing cooking knowledge from large-scale text corpora and then transferring the knowledge to the visual domain. This approach relieves us of the burden and impracticality of providing annotations for a virtually unlimited number of categories (dishes) and sub-categories (instructional steps). Our work is the first to tackle the problem of procedural activity understanding in this form; prior works in recognition are severely limited in the number of categories and steps~\cite{alayrac2016unsupervised,kuehne2014language,rohrbach2012database}, while works in anticipation rely on strong supervision~\cite{abu2018will,lan2014hierarchical,zhou2015temporal}.

Our work's second novelty is that we do not work with a closed set of labels derived from word tags. Instead, we train with and also predict full sentences, \eg, \emph{`Cook the chicken wing until both sides are golden brown.'} vs. \emph{`cook chicken'}. This design choice makes the problem more challenging but also brings several advantages. First, it adds qualifiers and richness to the instruction since natural language conveys much more information than simple text labels~\cite{lin2015generating,zhou2018towards}. Secondly, it allows for anticipation of not only actions but also objects and attributes. Finally, as a byproduct, it facilitates data collection, as the number of class-based annotations grows exponentially with the number of actions, objects, and attributes, leading to very long-tailed distributions~\cite{Damen2018EPICKITCHENS}.

When transferring procedural knowledge from text recipes to videos, we need to ground the text domain with the video and vice versa. This requires video with temporally aligned captions; to the best of our knowledge, YouCookII~\cite{zhou2018towards} is the only dataset with such labels. However, YouCookII lacks diversity in the number of dishes (89 dishes) and therefore the number of possible recipe steps. As such, we collect and present our new~\emph{Tasty Videos dataset V2}, a diverse set of 4022 different cooking recipes\footnote{ Collected from the website \texttt{\url{https://tasty.co/}}} each accompanied by a video, ingredient list, and temporally aligned recipe steps. Video footage is taken from a fixed bird's-eye view and focuses almost exclusively on the cooking instructions, making it well-suited for procedural understanding. 

Our main contributions are summarized as follows: 
\begin{itemize}
\item We are the first to explore action anticipation in a zero-shot setting by generalizing knowledge from text corpora and transferring it to the visual domain.

\item We propose a modular hierarchical model for learning multi-step procedures with text and visual context. Our model generalizes cooking knowledge and predicts coherent and plausible instructions for multiple steps into the future. The rich natural language predictions score higher in NLP metrics than state-of-the-art video captioning methods applied directly to the (future) video. 
 
\item We present a new and highly diverse dataset of cooking recipes. The dataset is publicly available\footnote{Tasty Video Dataset \texttt{\url{https://cvml.comp.nus.edu.sg/tasty}}} and will be of interest to those working in procedural video understanding, action recognition and anticipation, as well as other multimodal research in video and text. 
\end{itemize}

A preliminary version of this paper was published in~\cite{sener2019zero}. The current paper extends the previous work's model by integrating a temporal segment proposal method to the video encoder and additional losses at the recipe encoder to improve convergence. We add experiments comparing against recipe generation networks~\cite{salvador2019inverse} and verify that our hierarchical architecture better generalizes to previously unseen dishes or recipes. Finally, we extend the dataset by $60\%$ from 2511 to 4022 videos.

\section{Related Works}\label{relWorks} 

\subsection{Procedural Activity Modeling}
Understanding procedural activities and their sub-activities has been typically addressed as a supervised temporal video segmentation and recognition problem~\cite{rohrbach2012database,kuehne2014language,richard2016temporal}. A variety of models have been used, including conditional random fields~\cite{hoai2011joint}, hidden Markov models~\cite{lea2016segmental}, RNNs~\cite{singh2016multi} and recently temporal convolutional networks~\cite{lea2017temporal,farha2019ms}. Data and label-wise, these methods require video sequences in which every frame is labeled exhaustively, making it difficult to work at a large scale.

Alternative lines of work are either weakly supervised, using cues from accompanying narrations~\cite{alayrac2016unsupervised,malmaud2015s,sener2015unsupervised} or sub-activity orderings~\cite{huang2016connectionist,richard2017weakly,chang2019d3tw}, or are fully unsupervised~\cite{sener2018unsupervised,kukleva2019unsupervised}. Our work is similar to those using text cues; however, we do not rely on aligned video and text modalities for learning the activity models~\cite{alayrac2016unsupervised,sener2015unsupervised}. Assuming videos accompanied by temporally aligned narrations is not always the case for instructional videos as it is far more natural for people to talk about an action before/after performing it or to talk about an alternative action/object. Instead, we use a large corpus of unlabeled data in the text domain and use only a very small set of aligned data for grounding the visual evidence.

\subsection{Action Anticipation}
Action anticipation is the forecast of not-yet-observed actions into the future. Let $\tau_{\alpha}$ be the `anticipation time', \ie, how many seconds in advance to anticipate the next action. Then the task of action anticipation is to predict upcoming action, $\tau_{\alpha}$ seconds before it starts. In many recent works, $\tau_{\alpha}$ is considered to be 1 second~\cite{miech2019leveraging,furnari2019rulstm} but could vary between zero~\cite{lan2014hierarchical} and several seconds \cite{koppula2015anticipating}.

Early works in forecasting activities have been limited to simple movement primitives, such as \emph{reaching} and \emph{placing}~\cite{koppula2015anticipating}, or personal interactions like \emph{hand-shaking, hugging \etc}~\cite{lan2014hierarchical,vondrick2016anticipating}. More recent works anticipate up to thousands of action classes~\cite{Damen2018EPICKITCHENS} by defining actions as the composition of a verb and a noun. However, this does not scale as the number of actions grows, leading to a long-tail distribution. For example, in the new Epic-100 Dataset~\cite{Damen2020RESCALING} 
92\% of the actions are tail classes. 
Methods for anticipation include RNNs for encoding the observations~\cite{mahmud2017joint,furnari2019rulstm}, predicting future features and performing future action classification on these~\cite{gao2017red,zolfaghari2019learning}, employing knowledge distillation~\cite{tran2019back}, transferring knowledge from other sources such as word embeddings~\cite{camporese2020knowledge} or visual attribute classifiers~\cite{miech2019leveraging}. Moreover, recent interest in egocentric vision has started a new line of approaches dedicated to such recordings using gaze~\cite{shen2018egocentric} or hand-object interaction regions~\cite{liu2019forecasting,dessalene2020egocentric}.

Dense anticipation extends action anticipation to forecast multiple actions into the future. Examples include~\cite{abu2018will,Ke_2019_CVPR}, who propose two-stage methods that first perform action segmentation of the observed sequence before using the frame-wise labels as input for anticipation. The work of~\cite{sener2020temporal} bypasses the initial segmentation and performs dense anticipation in a single stage directly using a temporal aggregation framework. Our proposed approach can also predict actions multiple steps into the future, but unlike these methods, we do not work in a fully supervised framework. Furthermore, we do not require repetitions of activity sequences for training. Moreover, we are the first to predict the future in the form of sentences instead of category-based predictions as in ~\cite{abu2018will,Ke_2019_CVPR,sener2020temporal}. Similar to us, recent work~\cite{mahmud2020captioning} also predicts sentences for subsequent actions by extending their anticipation framework~\cite{mahmud2017joint}.

\subsection{Zero- and Few-Shot Learning in Video} 
Zero- and few-shot learning is more popular in the image domain and we refer the interested readers to two recent surveys~\cite{wang2020generalizing,wang2019survey}. Extensions to the video domain has been less explored. Early works on zero-shot learning on videos rely on attribute-oriented feature representations, which are then used to categorize the unseen videos~\cite{gan2016learning}. More recent works train temporal models that map video features to a semantic embedding space of categorical labels~\cite{zhu2018towards,hahn2019action2vec,brattoli2020rethinking} or sentence representations~\cite{zhang2018cross}. In a similar spirit to the embedding-based approaches, we ground the video to text by mapping video representations to the semantic space of step-wise instructions. Our approach extends the research on the zero-shot recognition of simple actions to the complex multi-step activities of procedural cooking videos. `Zero-shot' in our case refers to making predictions of previously unseen recipes. 

\subsection{Modeling Instructional Text in NLP} 
Cooking is a popular domain in NLP research since recipes are rich in natural language yet reasonably limited in scope. Cooking recipes are employed in tasks such as food recognition~\cite{Herranzfood}, recommender systems~\cite{Mindel2017} and indexing and retrieval~\cite{carvalho2018cross,salvador2017learning}. Modeling the procedural aspects of text and generating coherent recipes date back several decades~\cite{hammond1986chef,DaleRobert88}. Early works focus on parsing the recipes to extract verbs and ingredients~\cite{tenorth2010understanding,malmaud2014cooking,kiddon2015mise,jermsurawong2015predicting}. For example,~\cite{tenorth2010understanding} generate plans from textual instructions,~\cite{kiddon2015mise} map recipes to action graphs, and ~\cite{beetz2011robotic} use parsed instructions to make robots cook pancakes.

More recently, neural network-based solutions are popular, and they target especially improving the generated recipes' coherence. For example,~\cite{kiddon2016globally} train an encoder-decoder~\cite{sutskever2014sequence} with a checklist mechanism to keep track of the ingredients given as input. ~\cite{bosselut2018discourse} propose a reinforcement learning-based solution with discourse-aware rewards to encourage generating instructions in correct orders. ~\cite{h2020recipegpt} produce personalized recipes by fusing users' previously consumed recipes with an attention mechanism.

\subsection{Cooking Domain in Vision} 
In vision, cooking has been explored for procedural and fine-grained activity recognition~\cite{Damen2018EPICKITCHENS,kuehne2014language,rohrbach2012database,zhou2018towards}, 
temporal segmentation~\cite{kuehne2014language,zhou2018towards}, video-text alignment~\cite{malmaud2015s,lin2020recipe} and captioning~\cite{rohrbach2013translating,regneri2013grounding,zhou2018end}. There are several cooking and kitchen datasets~\cite{Damen2018EPICKITCHENS,malmaud2015s,sener2015unsupervised,kuehne2014language,zhou2018towards}; What's Cooking~\cite{malmaud2015s} and YouCookII~\cite{zhou2018towards} are the most similar to ours, featuring videos and accompanying recipe texts. YouCookII, however, has limited diversity with only 89 dishes; What's Cooking is large scale (180K recipes) but lacks temporal alignments between the video and recipe texts.

Some recent methods investigate learning image-text embeddings for image-based recipe retrieval. For example,~\cite{salvador2017learning} learn a joint embedding space of the recipes encoded with skip-thought vectors~\cite{sThought15} and associated food images using a pairwise ranking loss. This baseline is extended by learning with a triplet loss~\cite{carvalho2018cross} and hard sample mining~\cite{wang2019learning}. Instead of retrieval,~\cite{zhu2020cookgan} recently propose an image generation method from recipe text using an instruction encoder. 

An alternative application is to generate recipes from images.~\cite{salvador2019inverse} predicts ingredients of food images and use the ingredients as input for a transformer-based decoder. However,~\cite{salvador2019inverse} generates an entire recipe as one continuous text block, so recipes can only be as long as the allowed maximum length of the decoder (150 words).~\cite{wang2020decomposed} splits recipes into several chunks and predicts the instructions for each chunk guided by position encoders. 

\begin{figure*}[htb!]
\centering 
\includegraphics[width=1\textwidth]{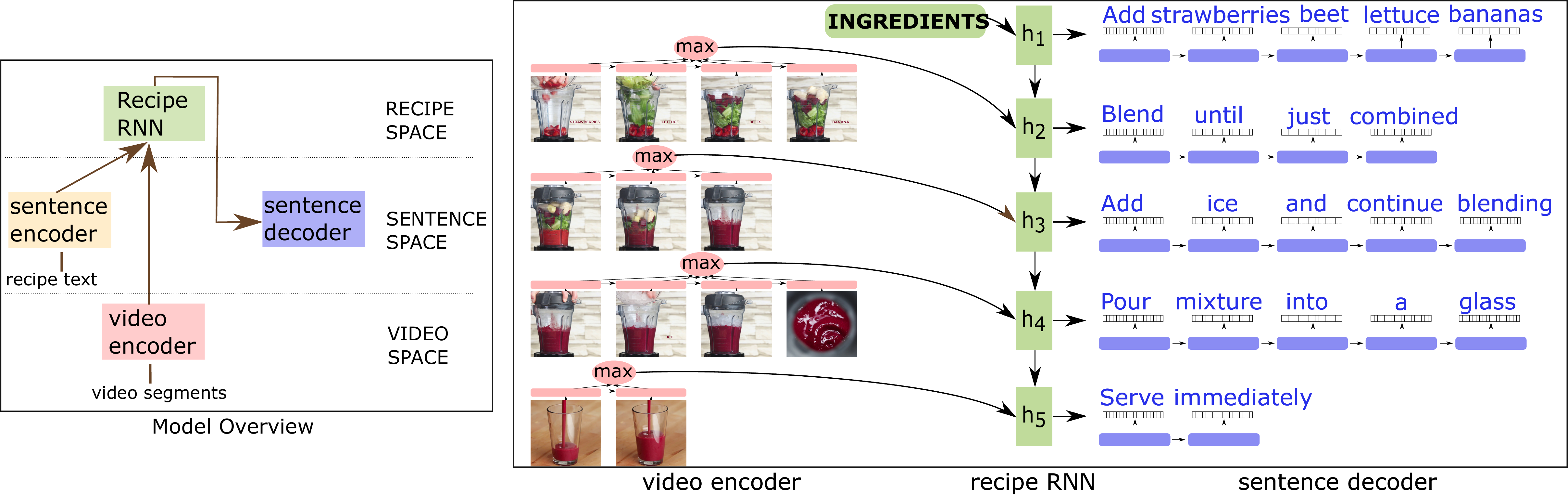}
\caption{
Our system is composed of four RNNs: a sentence encoder and a decoder, a video encoder, and a recipe RNN. Given the ingredients as initial input and context in either text or visual form, the recipe RNN recurrently predicts future steps. The sentence decoder converts predicted future steps back into natural language. We continue predicting future steps by repeatedly feeding the next steps encoded by the sentence or video encoder.
} 
\label{fig:overview} 
\end{figure*}

\section{Modeling Sequential Instructions}\label{sec:model}

Sequence-to-sequence learning~\cite{sutskever2014sequence} has made it possible to successfully generate continuous text and build dialogue systems~\cite{cho2014learning,vinyals2015neural}. Recurrent neural networks are used to learn rich representations of sentences~\cite{hill2016learning,Ba2016LayerN,sThought15} in an unsupervised manner, using the extensive amount of text that exists in book and web corpora. Examples include skip-thoughts vectors~\cite{sThought15} and FastSent~\cite{hill2016learning}, both of which are effective for generation tasks. However, for instructional text, such as cooking recipes, such representations do not fully capture the underlying sequential nature of the instruction set, and generations are not always coherent from one step to the next. As such, we propose a hierarchical model and dedicate two RNNs to represent the sentences and the steps of the recipe individually: the sentence encoder and the recipe RNN, respectively. A third RNN decodes predicted recipe steps back into sentence form for human-interpretable results (sentence decoder). These three RNNs are learned jointly as an auto-encoder in an initial training step. A fourth RNN encoding visual evidence (video encoder) is then learned in a subsequent step to replace the sentence encoder to enable interpretation and future prediction from video data. An overview is shown in Figure~\ref{fig:overview}, while details of the RNNs are given in Sections~\ref{sec:sentence} to~\ref{sec:videoenc}.
 
\subsection{Sentence Encoder and Decoder}\label{sec:sentence} 
The sentence encoder produces a fixed-length vector representation of each textual recipe step. We use a bi-directional LSTM, but rather than representing a sentence by the last step's hidden vector, we apply a (temporal) max-pooling over each dimension of the hidden units. This type of architecture and, in particular, the temporal pooling are shown to be successful in sentence encoding~\cite{conneauEMNLP2017}. More formally, let sentence $s_{j}$ from step $j$ of a recipe (we assume each step is one sentence) be represented by $M_j$ words, \ie, $s_{j} = \{w_{j}^{t}\}_{t=1...M}$ and ${\bm{x}}_{j}^t$ be the word embedding of word $w_{j}^{t}$. For each sentence $j$, at each (word) step $t$, the bi-directional LSTM based sentence encoder, $\text{SE}$, outputs $\bm{y}_{j}^t$: 

\begin{equation}\label{eq:sentenc} 
\bm{y}_{j}^t \! = \! \text{SE}( \left[\text{LSTM}\{\bm{x}_{j}^{1}, ..., \bm{x}_{j}^{t}\}, \text{LSTM}\{\bm{x}_{j}^{M_j}, ..., \bm{x}_{j}^{t}\}\right] ), 
\end{equation}

\noindent which is a concatenation of the hidden states from the forward and backward pass of the $\text{LSTM}$. The overall sentence representation $\bm{r}_{j}$ is determined by a dimension-independent max-pooling over the time steps, \ie,

\begin{equation}\label{eq:maxpool}
( \bm{r}_{j})_d = \max_{t \in \{1,...,M_j\} } (\bm{y}_{j}^t)_d,
\end{equation}

\noindent where $(\cdot)_d$, $d\!\in\!\{1, ..., D\}$, indicates the $d$-th element of the $D$-dimensional bi-directional LSTM outputs $\bm{y}_{j}^t$.

The sentence decoder, SD, is an LSTM-based neural language model that converts the fixed-length representation of the steps back into human-interpretable sentences. More specifically, given the vector prediction $\hat{\bm{r}}_{j}$ from the recipe RNN of step $j$, it decodes the sentence $\hat{s}_{j}$

\begin{equation}\label{eq:sentdec} 
 \hat{s}_{j} = \text{SD}( \text{LSTM}\{\hat{\bm{r}}_{j})\} = \{\hat{w}_{j}^1, ..., \hat{w}_{j}^{\hat{M_j}}\}.
\end{equation}
 
\subsection{Recipe RNN}\label{sec:recipeRNN}
We model the sequential ordering of recipe steps with a recipe encoder (RE), which is an LSTM that takes as input $\{\bm{r}_{j}\}_{j=1,...,N}$, \ie, fixed-length representations of the steps of a recipe with $N$ steps, where $j$ indicates the step index. At each recipe step, the hidden state of the RE $\bm{h}_{j}$ can be considered a fixed-length representation of all recipe steps $\{s_{1}, ..., s_{j}\}$ seen up to step $j$; we directly use this hidden state vector as a prediction for the sentence representation for step $j+1$, \ie,

\begin{equation}\label{eq:nextsent}
\hat{\bm{r}}_{j+1} = \bm{h}_{j} = \text{RE}(\text{LSTM}\{\bm{r}_{0},...,\bm{r}_{j}\}).
\end{equation}

\noindent The hidden state of the last step ${\bm{h}}_{N}$ can be considered as a representation of the entire recipe. Due to the standard recursion of the hidden states in LSTM, each hidden state vector, and therefore, each future step prediction, is conditioned on the previous steps. This allows predicting recipe steps that are plausible and coherent with respect to previous steps.

Recipes usually include an ingredient list, a rich source of information that can also serve as a strong modeling cue~\cite{kiddon2016globally,salvador2017learning,salvador2019inverse}. To incorporate the ingredients, we form an ingredient vector $\bm{I}$ for each recipe in the form of a one-hot encoding over a vocabulary of ingredients. $\bm{I}$ is then transformed with a separate fully connected layer in the recipe RNN to serve as the initial input, \ie, $\bm{r}_{0} = f(\bm{I})$. Note that our model is fully deterministic, so the same ingredient vector input will always lead to the same first instruction.
 
\subsection{Video Encoder}\label{sec:videoenc}

For inference, we would like the recipe RNN to interpret sentences from text and visual inputs. The modular nature of our model allows us to conveniently replace the sentence encoder with an analogous video encoder, VE. Suppose the $j^{\text{th}}$ video segment $c_{j}$ is composed of $C_j$ frames, \ie, $c_{j} = \{\bm{f}_{j}^t\}_{t=1,...,L}$~\footnote{We overload the word index $t$ from Eqs.~\ref{eq:sentenc} and~\ref{eq:maxpool} to also denote the frame index as the two are directly analogous in our encoders.}. Each frame $f_{j}^t$ is represented as a high-level CNN feature vector -- we use the last fully connected layer output of ResNet-50~\cite{he2016deep} before the softmax layer. Similar to the sentence encoding, $\bm{r}_{j}$, in Eqs.~\ref{eq:sentenc} and~\ref{eq:maxpool}, we determine the video encoding vector, $\bm{v}_{j}$, by applying a temporal max pooling over each dimension of the video segment representations, $\bm{z}_{j}^t$:

\begin{equation}\label{eq:maxpoolz}
( \bm{v}_{j})_d = \max_{t \in \{1,...,C_j\} } (\bm{z}_{j}^t)_d, \qquad \text{where}
\end{equation}

\begin{equation}\label{eq:videnc} 
 {\bm{z}_{j}^t} \! = \! \text{VE}(\left[\text{LSTM}\{\bm{f}_{j}^{1}, ..., \bm{f}_{j}^{t}\}\!, \text{LSTM}\!\{\bm{f}_{j}^{C_j}, ..., \bm{f}_{j}^{t}\}\right] ).
\end{equation}

\noindent The video encoder, $\text{VE}$, is trained such that $\bm{v}_{j}$ can directly replace $\bm{r}_{j}$. The inputs to our video encoder are frames from the video segments that correspond to individual recipe steps. We train our method on videos using ground truth segments for the video encoder. For testing, we use temporal segments either based on fixed temporal windows or from predicted segment proposals~\cite{zhou2018towards}.

\subsection{Model Learning}
Our full model is learned in two stages. First, the sentence encoder ($\text{SE}$), recipe RNN ($\text{RE}$) and sentence decoder ($\text{SD}$) are jointly trained end-to-end. Given a recipe of $N$ steps, we define our decoder loss, $ L_{d}$, as the negative log probability of each reconstructed word $t$ for each step $j$:

\begin{equation}\label{eq:objective_decoder}
 L_{d}(s_1, ..., s_{N}) = - \sum_{j=1}^{N} \sum_{t=1}^{M_j} \log P( w_{j}^{t} | w_{j}^{t'<t}, \hat{\bm{r}}_j),
\end{equation}

\noindent where $P( w_{j}^{t} | w_{j}^{t'<t}, \hat{\bm{r}}_j)$ is parameterised by a softmax function at the output layer of the sentence decoder to estimate the distribution over the words, $w$, in our vocabulary $V$. The overall objective is then summed over all recipes in the corpus. The loss is computed only when the LSTM is learning to decode a sentence. This first training stage is unsupervised, as the sentence encoder and decoder and the recipe RNN require only text inputs that can easily be scraped from the web without human annotations.

In a second step, we train the video encoder ($\text{VE}$) while keeping the recipe RNN ($\text{RE}$) and sentence decoder ($\text{SD}$) fixed. We simply replace the sentence encoder with the video encoder while applying the same loss function as defined in Eq.~\ref{eq:objective_decoder}. This step is supervised as it requires video segments of each step that are temporally aligned with the corresponding sentences.

In addition to the word-based reconstruction loss of Eq.~\ref{eq:objective_decoder}, we propose an L2 regularizer that encourages the predicted next step representation of $\hat{\mathbf{r}}$ of Eq.~\ref{eq:sentenc} to be faithful to the observed representation $\mathbf{r}$, \ie recipe loss $L_{r}$:
\begin{equation}\label{eq:objective_recipe}
L_{r} = \sum_{j=1}^{N} (\bm{r}_{j} - \hat{\bm{r}}_{j})^2, 
\end{equation}

\noindent We use the decoder loss, $L_{d}$, and recipe loss, $L_{r}$, together with weighting hyperparameter $\alpha$: 
\begin{equation}\label{eq:objective}
 L(s_1, ..., s_{N}) = L_{d} + \alpha L_{r},
\end{equation}
where $\hat{\bm{r}}_{j}$ is the predicted output for step j and ${\bm{r}}_{j}$ is the input from the sentence encoder for step j.

\subsection{Inference}\label{sec:inference}
During inference, we provide the ingredient vector $\bm{r}_{0}$ as an initial input to the recipe RNN, which then outputs the predicted vector $\hat{\bm{r}}_{1}$ for the first step of the video (see Figure~\ref{fig:overview}). We use the sentence decoder and generate the corresponding first sentence, $\hat{s}_{1}$. Then, we sample a sequence of frames from the video and apply the video encoder to generate $\bm{v}_{1}$, which we again provide as an input to the recipe RNN. The output prediction of the recipe RNN, $\hat{\bm{r}}_{2}$, is for the second step of the video. We again use the sentence decoder and generate the corresponding sentence $\hat{s}_{2}$.

Our model is not limited to one-step-ahead predictions: for further predictions, we can simply apply the predicted output $\hat{\bm{r}}_{j}$ as contextual input ${\bm{r}}_{j}$. During training, instead of always feeding in the ground truth ${\bm{r}}_{j}$, we sometimes (with 0.5 probability after the 5th epoch) use our predictions, $\hat{\bm{r}}_{j}$, as the input for the next-step predictions, which helps us with being robust to feeding in bad predictions~\cite{bengio2015scheduled}.

\begin{figure*}[htb!] 
\centering 
\subfloat[Number of steps]{\label{fig:mdleft}{\includegraphics[width=0.24\textwidth]{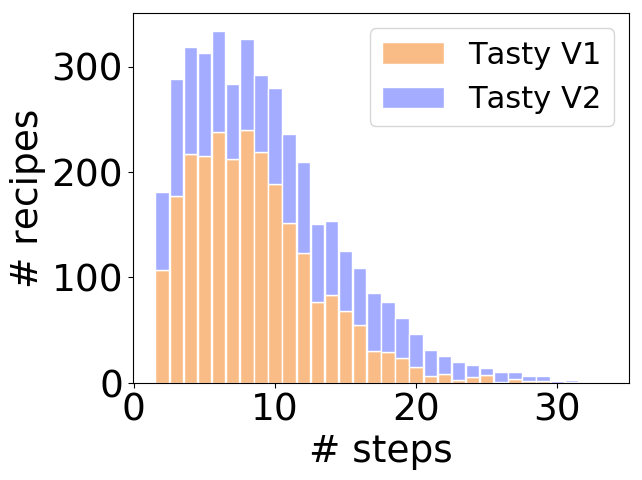}}}\hfill 
\subfloat[Number of ingredients]{\label{fig:mdleft}{\includegraphics[width=0.24\textwidth]{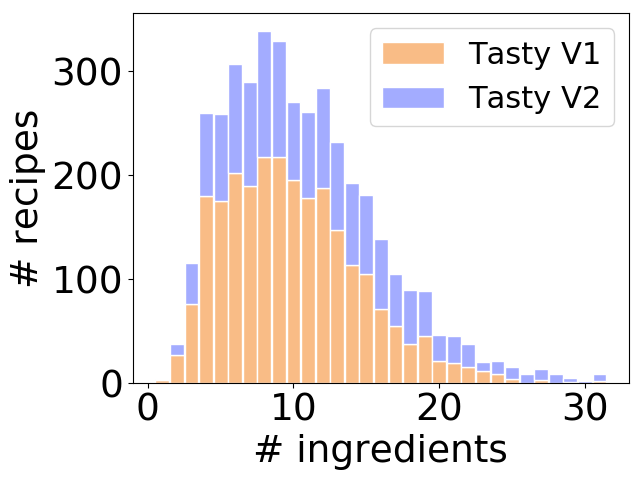}}}\hfill 
\subfloat[Step duration (seconds) ]{\label{fig:mdleft}{\includegraphics[width=0.25\textwidth]{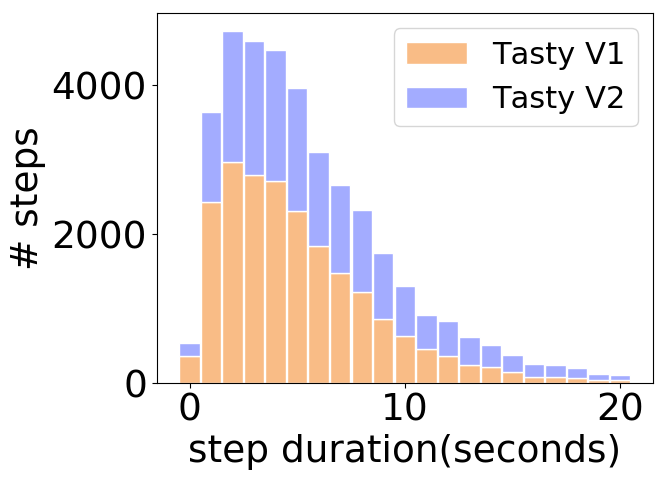}}}\hfill 
\subfloat[Video duration (seconds)]{\label{fig:mdleft}{\includegraphics[width=0.23\textwidth]{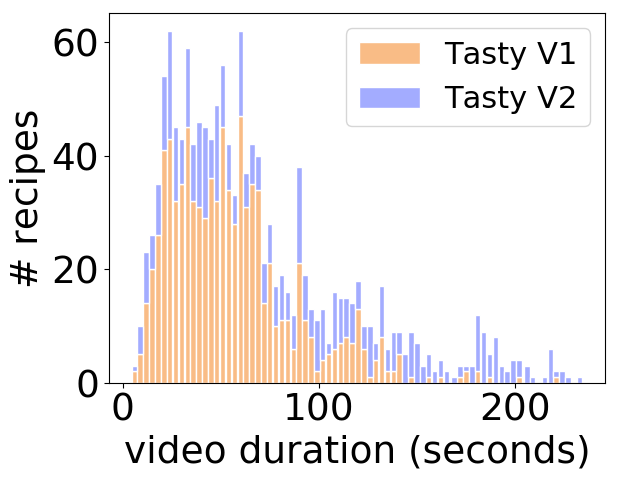}}}\hfill 
\caption{Dataset distributions.} 
\label{fig:tasty_stat_steps} 
\end{figure*}

\begin{table*}[!ht]
\centering
\resizebox{\columnwidth}{!}{
\setlength{\tabcolsep}{10.7pt}
\begin{tabular}{@{}llllllcl@{}}
\toprule
 Tasty & \#videos &\begin{tabular}[c]{@{}l@{}} \#images / \\ frames\end{tabular} & \begin{tabular}[c]{@{}l@{}}\#unique\\ ingredients\end{tabular} & \begin{tabular}[c]{@{}l@{}}avg. \#steps / \\ avg. \#segments\end{tabular} & \begin{tabular}[c]{@{}l@{}}\#steps / \\ \#segments\end{tabular} & \begin{tabular}[c]{@{}l@{}}avg. segment\\ duration\end{tabular} & \begin{tabular}[c]{@{}l@{}}avg. video\\ duration\end{tabular} \\ \midrule
 V1 & 2511 & 4.1M & 1228 & 9 & 21243 & 5s & 54s \\
V2 & 4022 & 8.6M & 1542 & 10 & 37530 & 6s & 71s \\ \midrule
Recipe1M\cite{salvador2017learning} & - & 887K & 3769 & 9 & - & - & - \\ \midrule
YouCookII~\cite{zhou2018towards} & 2000 & 15.8M & 828 & 8 & 15400 & 19.6s & 303s \\ 
Epic-Kitchens\cite{Damen2018EPICKITCHENS} & 432 & 11.5M & - & - & 39596 & 1.9s & 426s \\ \bottomrule
\end{tabular}} 
\caption{
Comparisons of our Tasty V1 and V2 datasets with relevant datasets. Recipe1M includes textual recipes and recipe images, while YouCookII and Epic-Kitchens are video-based datasets. 
} 
\label{tab:tasty_stats}
\end{table*}

\subsection{Implementation and Training Details} 
We use a vocabulary $V$ of 30171 words provided by the Recipe1M dataset~\cite{salvador2017learning}; words are represented by a 256-dimensional embedding shared by the sentence encoder and decoder. We use the ingredient vocabulary from the training set of Recipe1M; the one-hot ingredient encodings are mapped into a 1024 dimensional vector $\bm{r}_0$. The RNNs are all single-layer LSTMs implemented in PyTorch; $\text{SE}$, $\text{VE}$, $\text{SD}$ have 512 hidden units, while $\text{RE}$ has 1024. We train our model using the Adam optimizer~\cite{kingma2014adam} with a batch size of 50 recipes and a learning rate of 0.001. We train our text-based model ($\text{SE}$, $\text{RE}$, $\text{SD}$) for 50 epochs and the visual model ($\text{VE}$, $\text{RE}$, $\text{SD}$) for 25 epochs. We use $\alpha =0.1$ for $ L_{r}$. The text-based model trained with $L_{r}$ converges faster, so we train this variant for only 10 epochs.

\section{Tasty Videos Dataset V1 and V2}\label{sec:tastydataset}
In our original publication \cite{sener2019zero}, we released the \emph{Tasty Videos Dataset} with 2511 unique recipes, which we will refer to as $V1$. We have since extended the dataset to 4022 unique recipes -- \emph{Tasty Videos Dataset V2}. All text recipes and videos are collected from Buzzfeed's Tasty website~\footnote{\texttt{\small\url{https://tasty.co}}}. We make publicly available the links to each recipe page, computed features, and temporal annotations~\footnote{\texttt{\small\url{https://cvml.comp.nus.edu.sg/tasty}}} for both versions of the dataset. 

In our dataset, each recipe has an ingredient list, step-wise instructions, and a video demonstrating the preparation of the dish. The videos in this dataset are captured with a fixed overhead camera and focus entirely on preparing the dish (see Figure \ref{fig:overview}). This viewpoint removes the added challenge of distractors and irrelevant actions. This simplification is not reflective of in-the-wild environments but it does allow us to focus our scope on modeling the sequential nature of instructional videos, which is already a highly challenging and open research topic. The videos are designed to be sufficiently informative visually without the need for any narrations.
 
Tasty V2 features 400 test, 400 validation and 3222 training instances. When creating the splits of both Tasty V1 and V2, we grouped the recipes based on their similarities, \eg muffins, tarts, and pizza, and then split the recipes in each category into training, validation, and testing sets. 
The recipes in the test set can be further split into those with similarities in the training set, \eg "strawberry pretzel cheesecake" vs. "carrot cake cheesecake" and those without similarities, \eg "okonomiyaki". Tasty V1 test set has 183 recipes with similarities to the training and 72 without. Tasty V2 has more recipes overall, so there are more test recipes with similarities; 366 recipes are with similarities and 34 are without.
 
As the number of recipes increase, the overlapping steps across recipes also increase. To capture the extent of overlap, we create exhaustive verb and ingredient pairs for each sentence. 34\% of the pairs are unseen during evaluation in Tasty V1, and 24\% are in the larger Tasty V2. If we simply concatenate the verbs and ingredients in each sentence to represent the sentences, then 64\% and 62\% of the tuples are unseen during evaluation in Tasty V1 and V2, respectively.

Other datasets feature crowd-sourced text~\cite{zhou2018towards,Damen2018EPICKITCHENS}; the recipes in our dataset are written by experts. This ensures specificity and richness in the instruction. For each recipe step, which corresponds to a single sentence, we annotate the temporal boundaries of the step in the video. We omit annotating steps without visual correspondences, such as alternative recommendations, non-visualized instructions like \emph{`Preheat oven.'}\, and stylistic statements such as \emph{`Enjoy!'}. For both Tasty V1 and V2, we define a split ratio of 8:1:1 for the training, validation, and testing sets.

We present the statistics of our datasets in Table~\ref{tab:tasty_stats} and in Figure~\ref{fig:tasty_stat_steps}. We compare our dataset to the relevant datasets of Recipe1M and YouCookII as well a large-scale activity dataset Epic-Kitchens in Table~\ref{tab:tasty_stats}. Recipe1M is a large-scale dataset with, as the name suggests, approximately one million recipes with a recipe name, list of ingredients, a sequence of instructions, and images of the final dish for each recipe. YouCookII is a collection of cooking videos from YouTube with around 2000 videos of 89 dishes. Videos are captured from a third-person viewpoint. Each dish has an average of 22 videos, each with an average of 8 steps. The videos are annotated with the temporal boundaries of each step and their corresponding descriptions. Epic-Kitchens is a large-scale egocentric activity dataset with 39K action segments with category-based labels and is frequently used for action recognition and anticipation. 

Tasty V2 extends the number of visual segments by 76\%. It includes around 37K segments that correspond to single sentence recipe steps. This number is comparable to the large-scale Epic-Kitchens dataset, which contains 39K segments, and is higher than the number of segments in YouCookII with its 15K segments. Compared to the 1M recipes of Recipe1M, our dataset with 4022 recipes covers 40\% of the 3769 ingredients in Recipe1M. Compared to YouCookII, our dataset includes a diverse list of dishes (4022 vs. 89) and ingredients (1542 vs. 828). One notable difference between these datasets is the segment/step granularity, which is on the order of a few seconds in our dataset and Epic but is coarser for YouCookII (19.6 seconds on average). The videos in our dataset are short (on average 54/71 seconds for V1/V2) yet contain a challenging number of steps (on average 9/10 for V1/V2). The recipes in YouCookII and Recipe1M have a similar number of steps (9/8 respectively). 

We note that the Tasty dataset also has the potential for tasks beyond anticipation, such as temporal action segmentation, dense video captioning, and object state recognition. In this paper, we compute several video captioning baselines, and we encourage the community to further develop models to tackle our challenging zero-shot dataset.

\begin{figure*}[t] 
\footnotesize
\bgroup
\def\arraystretch{1.30}
\setlength\tabcolsep{0.3em}
\setstretch{0.80}
\begin{tabularx}{1.00\textwidth}{|c|X|X|c|c|c|c|c|}
\hline
& \textit{Ground Truth (GT)} & \textit{Prediction} & \tiny{\textit{BLEU1}} & \tiny{\textit{BLEU4}} & \tiny{\textit{MET.}} & \tiny{\textit{HUM1}}& \tiny{\textit{HUM2}}\\\hline
\textit{ING} & bacon, brown sugar, cooking spray, breadsticks & & & & & & \\\hline 
\textit{step1} & Preheat oven to 325 degrees F ( 165 degrees C ). & Preheat oven to 400 degrees F. & \cellcolor{orange!36}36.0 & \cellcolor{red!0}0.0 & \cellcolor{blue!26} 26.0 & \cellcolor{greenff!50} 1.5 & \cellcolor{greenff!50} 1.5 \\\hline
\textit{step2} & Line 2 baking sheets with aluminum foil or parchment paper and spray with cooking spray.	& Line a baking sheet with aluminum foil. & \cellcolor{orange!23}23.0 & \cellcolor{red!0}0.0 & \cellcolor{blue!23} 23.0 & \cellcolor{greenff!25}1.0 & \cellcolor{greenff!25}1.0\\\hline
\textit{step3} & Wrap 1 bacon strip around each breadstick, leaving about 1 inch uncovered on each end. & Place bacon strips in a single layer on the prepared baking sheet.&\cellcolor{orange!13} 13.0 & \cellcolor{red!0}0.0 & \cellcolor{blue!9}9.0 & \cellcolor{greenff!10}0.5 & \cellcolor{greenff!50} 1.5 \\\hline
\textit{step4} & Place wrapped breadsticks on the prepared baking sheet. & Place rolls on a baking sheet. & \cellcolor{orange!48}48.0 & \cellcolor{red!0}0.0 &\cellcolor{blue!30}30.0 & \cellcolor{greenff!50} 1.5 & \cellcolor{greenff!50} 1.5 \\\hline
\cellcolor{lime!8}\textit{step5} & \cellcolor{lime!8}Sprinkle brown sugar evenly over breadsticks. &\cellcolor{lime!8} Bake in the preheated oven until breadsticks are golden brown, about 15 minutes. & \cellcolor{orange!15}15.0 & \cellcolor{red!0}0.0 &\cellcolor{blue!13}13.0 & \cellcolor{greenff!0}0.0 & \cellcolor{greenff!50} 1.5\\\hline
\cellcolor{blue!5}\textit{step6} & \cellcolor{blue!5}Bake in the preheated oven until bacon is crisp and browned, 50 to 60 minutes. & \cellcolor{blue!5}Bake in preheated oven until bacon is crisp and breadsticks are golden brown, about 15 minutes.& \cellcolor{orange!63}63.0 & \cellcolor{red!43}43.0 &\cellcolor{blue!36}36.0 & \cellcolor{greenff!25}1.0 & \cellcolor{greenff!25}1.0\\\hline 
\textit{step7} & Cool breadsticks on a piece of parchment paper or waxed paper sprayed with cooking spray. & Remove from oven and let cool for 5 minutes.&\cellcolor{orange!6} 6.0 & \cellcolor{red!0}0.0 &\cellcolor{blue!4}4.0 & \cellcolor{greenff!10}0.5 & \cellcolor{greenff!50} 1.5 \\\hline
\end{tabularx}
\egroup 
\caption{
Predictions of our text-based method for \emph{`Candied Bacon Sticks'} along with the automated scores and human ratings. For \emph{`HUMAN1 (HUM1)'} we asked the raters to directly assess how well the predicted steps match the corresponding Ground Truth~(GT) sentences; for \emph{`HUMAN2 (HUM2)'}, we asked them to judge if the predicted step is still a plausible future prediction (see Sec.~\ref{sec:human_study}). Our prediction for step6 matches the GT well, while that for step5 does not. However, according to \emph{`HUMAN2 (HUM2)'} score, our step5 prediction is still a plausible future action.
} 
\label{fig:example_preds} 
\end{figure*}

\section{Experiments: Text}\label{sec:experiments_prev}

\subsection{Datasets and Evaluation Measures}
We experiment with Recipe1M~\cite{salvador2017learning}, YouCookII~\cite{zhou2018towards}, and Tasty Videos V1 and V2.

We use the ingredients and instructions from the training split of the Recipe1M dataset to learn our sentence encoder (Eq.~\ref{eq:sentenc}), sentence decoder (Eq.~\ref{eq:sentdec}), and recipe RNN (Eq.~\ref{eq:nextsent}). To learn the video encoder (Eq.~\ref{eq:videnc}), we use the aligned instructions and video data from the training split of either YouCookII or Tasty datasets. We evaluate our model's prediction capabilities with text inputs from Recipe1M and video and text inputs from YouCookII and Tasty Videos.

Our predictions are in sentence form; evaluating the quality of generated sentences is known to be difficult in captioning and natural language generation ~\cite{vedantam2015cider,lopez2008statistical}. We apply a variety of measures to offer a broad assessment. First, we target the matching of ingredients and verb keywords since they indicate the next active objects and actions and are analogous to the assessments of action anticipation ~\cite{Damen2018EPICKITCHENS}. Second, we evaluate using sentence-matching scores BLEU (BiLingual Evaluation Understudy) ~\cite{papineni2002bleu} and METEOR (Metric for Evaluation of Translation with Explicit ORdering) ~\cite{banerjee2005meteor}, which are also used for video captioning methods ~\cite{regneri2013grounding,rohrbach2013translating,zhou2018end}. BLEU computes an n-gram-based precision for predicted sentences \emph{w.r.t.} the ground truth sentences. METEOR creates an alignment between the ground truth and predicted sentence using the exact word matches, stems, synonyms, and paraphrases; then, it computes a weighted F-score with an alignment fragmentation penalty.

For the uninformed reader, sentence scores like BLEU and METEOR are best at indicating sentences with precise word matches to the ground truth (GT). There are variations between sentences conveying the same idea in natural language, so automated scores may fail to match sentences a human would consider equivalent. This is true even for text with very specific language, such as cooking recipes. For example, for the ground truth sentence \emph{`Garnish with the remaining wasabi and sliced green onions.'}, our method predicts \emph{`Transfer to a serving bowl and garnish with reserved scallions.'}. For a human reader, this is half correct, since \emph{`scallions'} and \emph{`green onions'} are synonyms, yet this example would have only a BLEU1 score of $30.0$, BLEU4 of $0.0$ and METEOR of $11.00$. For another, for the ground truth sentence \textit{`Place patties on the grill, and cook for 5 minutes per side.'} versus a prediction by our model \textit{`Place on the grill, and cook for about 10 minutes, turning once.'}, we would have a BLEU1 score of 65.0, BLEU4 of 44.0, and METEOR of 29.0. In this regard, we note that BLEU and METEOR scores offer only a limited ability to evaluate the predicted sentences.

The gold standard to evaluate dialogue generation~\cite{LiuLSNCP16} and captioning~\cite{li2018jointly} is human subject ratings. Therefore, we conduct a user study and ask people to assess how well the predicted step matches the ground truth in meaning; if it does not match, we ask if the prediction would be plausible for future steps. This gives flexibility in case predictions do not follow the exact aligned order of the ground truth, \eg, due to missing steps not being predicted or steps that are slightly out of order (see Figures ~\ref{fig:example_preds}, ~\ref{fig:overview_example2} and ~\ref{fig:overview_example}).

\begin{figure*}[!htb]
\centering
 \begin{tabular}{@{}c@{}}
 \includegraphics[width=0.57\textwidth]{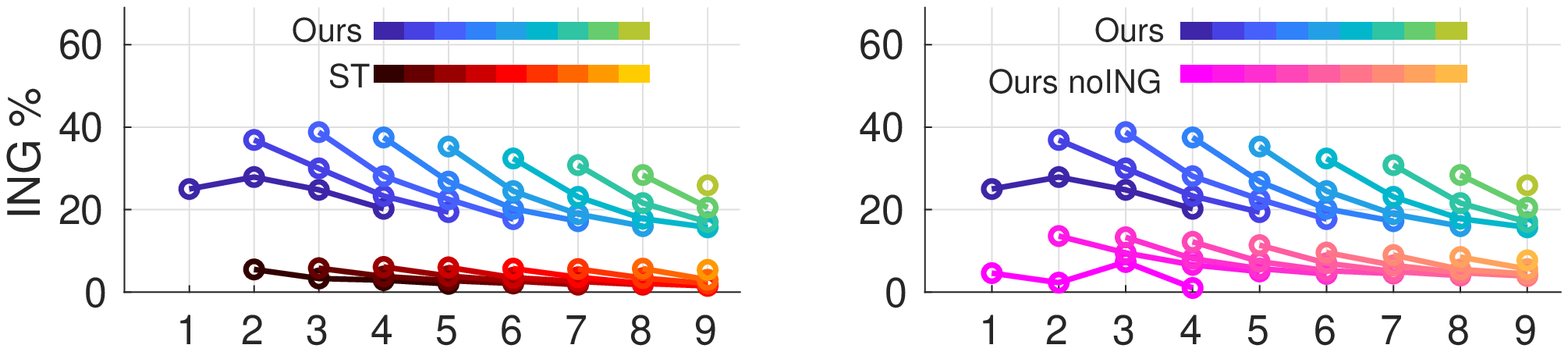}\\ 
 \small (a) Ingredient recall.
 \end{tabular}
 \hspace{3mm}
 \begin{tabular}{@{}c@{}}
 \includegraphics[width=0.30\textwidth]{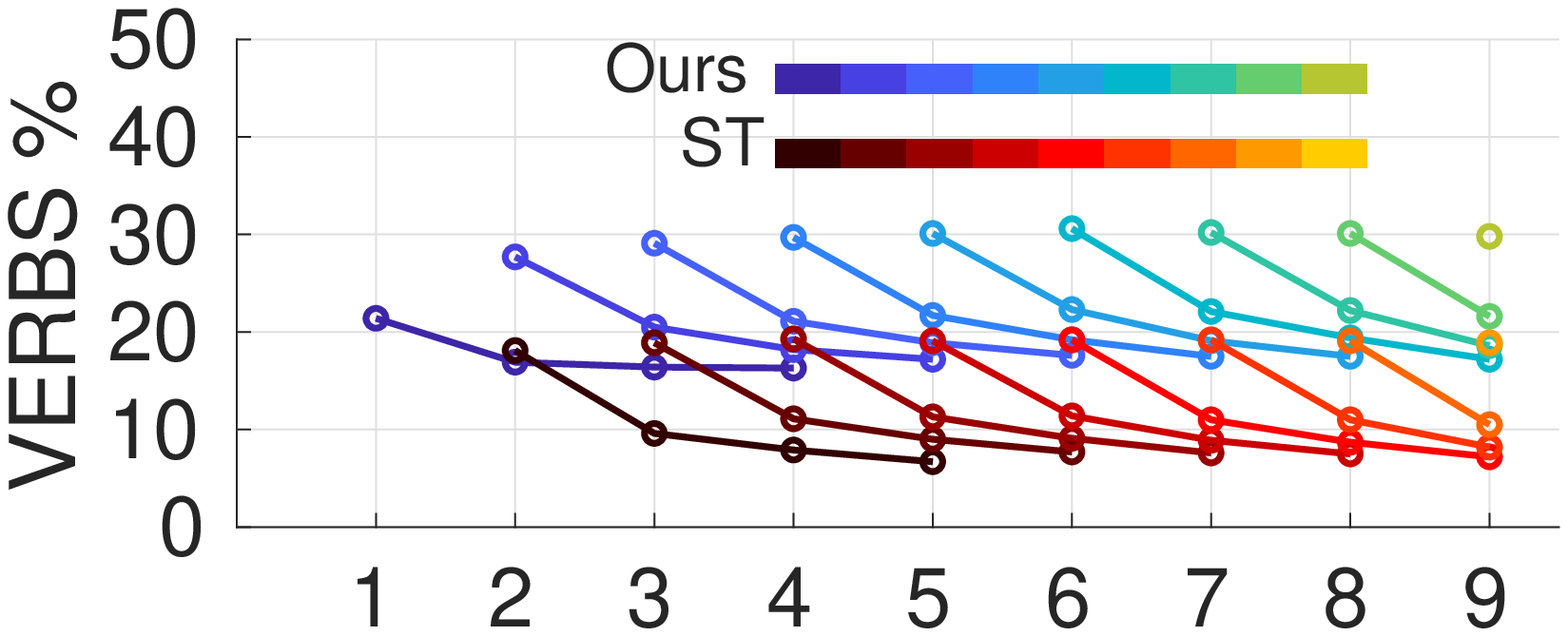}\\ 
 \small (b) Verb recall.
 \vspace{3mm}
 \end{tabular}
 \begin{tabular}{@{}c@{}}
 \includegraphics[width=0.89\textwidth]{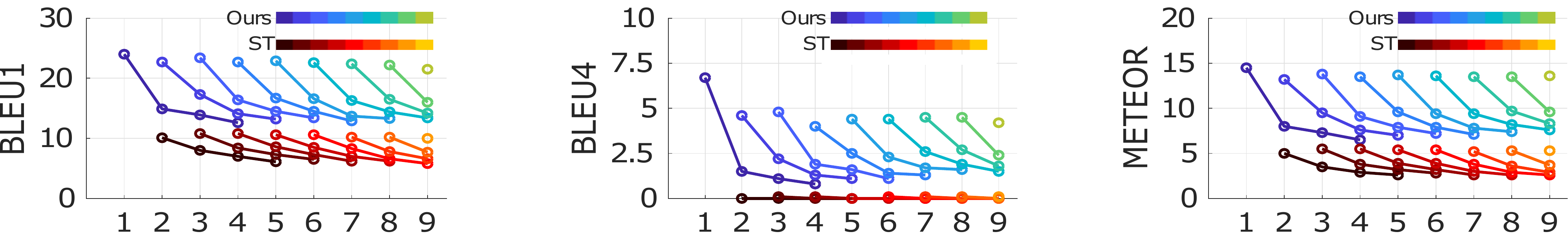}\\ 
 \small (c) Sentence scores.
 \end{tabular}
\caption{
Comparisons on Recipe1M's test set. (a) recall of ingredients predicted by our model (\emph{`Ours'}), skip-thought vectors (\emph{`ST'}), and our model trained without ingredients (\emph{`Ours noING'}). (b) verb recall of our model (\emph{`Ours'}) vs \emph{`ST'}. (c) BLEU1, BLEU4, METEOR scores for our model (\emph{`Ours'}) vs. \emph{`ST'}. The x-axes in the plots indicate the step number being predicted in the recipe; each curve begins on the first prediction, \ie, the $(j+1){\text{th}}$ step after having received steps $1$ to $j$ as input. }
\label{fig:all_scores_recipe1m}
\end{figure*}

\begin{figure*}[!htb]
\centering 
 \includegraphics[width=1\textwidth]{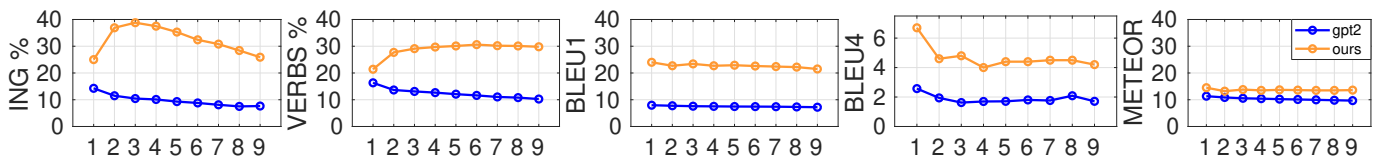}
\caption{Comparisons with GPT-2~\cite{radford2019language} on Recipe1M's test set. Compared to our text-based model, GPT-2 has lower performance in all scores, indicating the importance of a dedicated hierarchical model like ours.}
\label{fig:all_scores_recipe1m_gpt}
\end{figure*}

\subsection{Learning of Procedural Knowledge} 
We first verify the learning of procedural knowledge with a text-only model, \ie, the sentence encoder, sentence decoder, and the recipe RNN, by evaluating on Recipe1M's test set of 51K recipes. For a recipe of $N$ steps, we evaluate our model's ability to predict steps $j\!+\!1$ to $N$, conditioning on steps 1 to $j$ as input context. For comparison, we look at the generations from the commonly used sequence-to-sequence model skip-thought (ST) vectors~\cite{sThought15}. Skip-thought vectors are trained to decode temporally adjacent sentences from a current encoding, \ie, given step $j$ to the encoder, the decoder predicts steps $j\!+\!1$ and $j\!-\!1$, and have been shown to be successful in generating continuous text~\cite{cho2014learning,vinyals2015neural,kiddon2016globally}.

We train the skip-thought vectors on the training set of the Recipe1M dataset; because the model is not designed to accept an ingredient list as a $0^{\text{th}}$ or initialization step, we make skip-thought predictions only from the second step onwards. We report our results for the recipes in the entire test set of the Recipe1M dataset in Figure~\ref{fig:all_scores_recipe1m} (a), (b) and (c). We report scores of the predicted steps averaged over multiple recipes. Only those recipes that have at least $j$ steps contribute to the average for step $j$.

\subsubsection{Key Ingredients} 
We first look at our model's ability to predict ingredients and verbs on the Recipe1M dataset. For ingredients, we also compare with a variant of our model without any ingredients (\emph{`Ours noING'}) where we train our network without ingredient inputs. To evaluate recall, we do not directly cross-reference the ingredient list but instead limit the evaluation to ingredients mentioned explicitly in the recipe steps. This is necessary to avoid ambiguities that may arise from specific instructions such as \emph{`add chicken, onion, and bell pepper'} versus the more vague \emph{`add remaining ingredients'}. Furthermore, the ingredient lists in Recipe1M are often automatically generated and may be incomplete.

In Figure~\ref{fig:all_scores_recipe1m}(a), we compare the recall of the ingredients detected in our predicted steps versus steps generated by skip-thought vectors and our model trained without ingredient inputs. We can see that our model's predictions successfully incorporate relevant ingredients with recall rates as high as 39.6\% with the predicted next step, 31.0\% with the second, 24.8\% with the third and 20.2\% with the predicted fourth step. The overall recall decreases with the later steps. This is likely due to the increased difficulty once the overall number of ingredient occurrences decreases, which tends to happen in later steps. Based on the ground truth, we observe that the majority of the ingredients occur in the early and middle steps and decrease in the last steps. The last steps are usually related to the already completed dish and do not explicitly mention as many ingredients as the earlier steps.

Compared to skip-thought vectors, our predictions' ingredient recall is higher regardless of whether or not ingredients are provided as an initial input. Without ingredient input, the overall recall is lower, but after the initial step, our model's recall increases sharply, \ie, once it receives some context. Our model without the ingredient input still performs better than the skip-thought predictions. We attribute this to the strength of our model to generalize across related recipes so that it is able to predict relevant co-occurring ingredients. Our predictions include common ingredients such as \emph{salt, butter, eggs} and \emph{water} and also recipe-specific ones such as \emph{couscous, zucchini}, or \emph{chocolate chips}. While skip-thought vectors predict some common ingredients, they fail to predict recipe-specific ingredients.

\begin{figure*}[htb]
\centering 
	\includegraphics[width=0.99\textwidth]{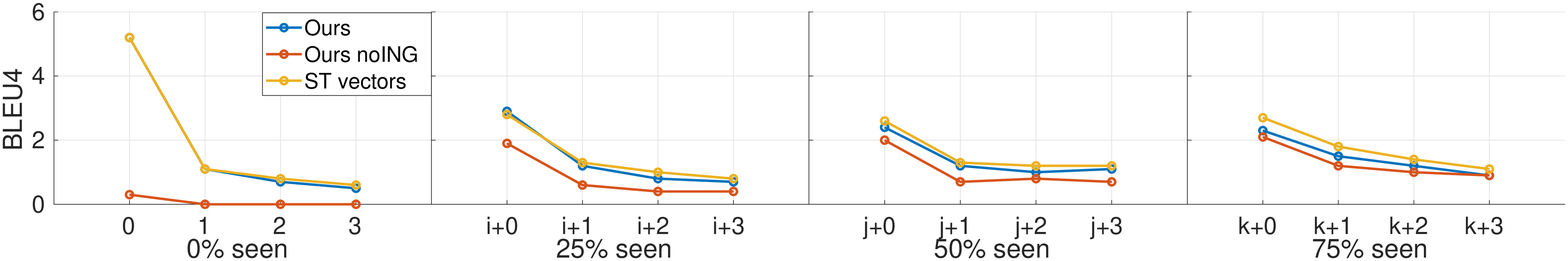} 
	\includegraphics[width=0.99\textwidth]{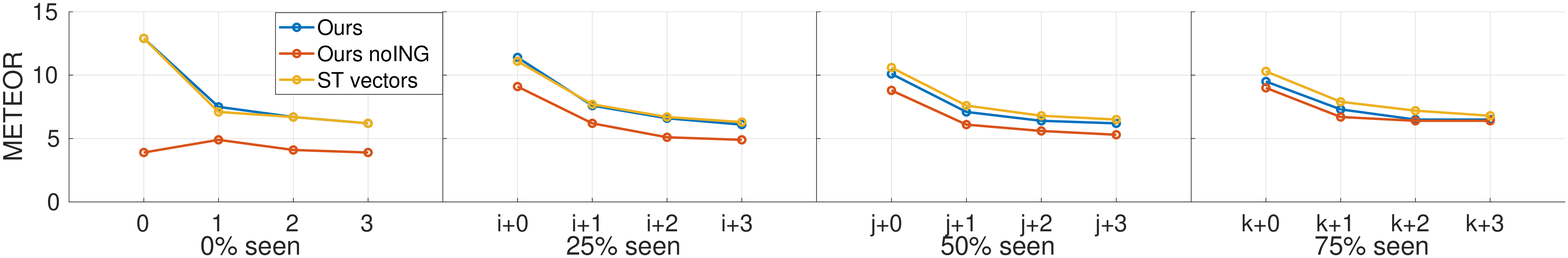} 
\caption{
Ablations on the interchangeability of the sentence encoder and the influence of ingredient inputs, evaluated on Recipe1M's test set. We compare the sentence scores of our joint model (\emph{`Ours'}), our joint model without ingredient inputs (\emph{`Ours noING'}), and our model where the sentence encoder is replaced with pre-trained skip-thought vectors (\emph{`ST vectors'}). \emph{`$X\% $ seen'} refers to the number of steps the model receives as input, while predicting the remaining $(100 - X) \% $.
}
\label{fig:im2recipe_percentage} 
\end{figure*} 

\subsubsection{Key Verbs} 
Key verbs indicate the main action for a step and are also cues for future steps both immediate (\eg, \emph{`mix'} after \emph{`adding'} ingredients into a bowl) and long-term (\eg, \emph{`bake'} after \emph{`preheating'} the oven). We tag the verbs in the training recipes with a natural language toolkit~\cite{nltk} and select the 250 most frequent verbs for evaluation. Similar to ingredients, we check the recall for only those verbs appearing in the ground truth steps. In the ground truth steps, there are between 1.55 and 1.85 verbs per step, \ie, steps often include multiple verbs such as \emph{`add and mix'}.

Figure~\ref{fig:all_scores_recipe1m} shows that our model recalls up to 30.6\% of the verbs with the predicted next step(b). Our model's performance is poor in the first steps, due to ambiguities when given only the ingredients without any further knowledge of the recipe. After the first steps, our model's performance quickly increases and stays consistent across the remaining steps. In comparison, the ST model's best recall is only 20.1\% for the next-step prediction.

\subsubsection{Sentences} 
Key ingredients and verbs alone do not capture the rich instructional nature of recipe steps; compare, \eg, \emph{`whisk'} and \emph{`egg'} to \emph{`Whisk the eggs till light and fluffy'}. As such, we also evaluate the quality of the entire predicted sentences based on the BLEU1, BLEU4, and METEOR scores. We compare with skip-thought vectors in Figure~\ref{fig:all_scores_recipe1m}~(c). Our results for the BLEU1 scores are consistently high, at around 22.7 for the next-step predictions, with a slight decrease towards the end of the recipe. Predictions further than the next step have lower scores, though they stay above 12.0. The BLEU4 scores are highest in the very first step, around 5.8, and range between 0.7 and 4.3 over the remaining steps. The high performance in the beginning is because many of the recipes start with common instructions (\eg, \emph{`Preheat oven to X degrees'}, \emph{`In a large skillet, heat the oil'}). For similar reasons, we also do well towards the end of recipes, where instructions for serving and garnishing are common (\eg, \emph{`Season with salt and pepper'}). Trends for the METEOR score are similar to our BLEU1 scores. METEOR scores are above 13.0 for the next-step predictions and do not go lower than 6.5 for the further-step predictions.
 
Our proposed method outperforms skip-thought predictions across the board. In fact, predictions up to four steps into the future surpass the predictions made by skip-thought predictions only one step ahead. This can be attributed to the dedicated long-term modeling of the recipe RNN; as such, we are able to incorporate the context from all sentence inputs up until the present. In contrast, skip-thoughts are Markovian in nature and can only take the current step into account.

Our model predicts coherent and plausible instructional sentences, as shown in Figure~\ref{fig:example_preds}. One interesting and unexpected outcome of our model is that it also makes recommendations.
In cooking recipes, one finds not only strict recipe steps but also suggestions based on the writer's experience (\eg, \emph{`If using wooden skewers, make sure to soak in water.'}).
Our learned model also generates such suggestions. For example, for the ground truth \emph{`If it's too loose at this point, place it in the freezer for a little while to let freeze.'}, our model predicts \emph{`If you freeze it, it will be easier to eat'}. 
 
We also compare our model with a transformed based architecture, GPT-2~\cite{radford2019language}, in Fig.~\ref{fig:all_scores_recipe1m_gpt}. We observe a similar trend for (ST) vectors~\cite{sThought15} in all scores, indicating a significant gap and highlighting the importance of a dedicated hierarchical model like ours.

\subsection{Encoder Modularity}

Since our network is modular, we check the interchangeability of the sentence encoder by replacing it with skip-thought vectors trained on the Recipe1M dataset, as provided by \cite{salvador2017learning}. For this experiment, we train the recipe RNN and sentence decoder jointly using the pre-trained skip-thought vectors as sentence representations. The recipe RNN and sentence decoder have been trained with the same parameter settings as our full model in all the ablation studies.

Figure~\ref{fig:im2recipe_percentage} compares sentence scores of our joint model (\emph{`Ours'}), our joint model trained without ingredient inputs (\emph{`Ours noING'}), and our model using the pre-trained skip-thought vectors (\emph{`ST vectors'}) when \emph{`$X\%$'} where $X =\{0, 25, 50, 75\}$, of a recipe is observed. Our sentence encoder performs on par with skip-thought vectors as an encoding. 
An advantage of our model, however, is that our encoder and decoder can be trained jointly and there is no need for a separate pre-training of a sentence auto-encoder, as required when using skip-thought vectors as input. Similar to our observations for ingredient recall, we see that ingredient information is very important for predicting sentences, especially for the initial steps.
In subsequent steps, when 25\%, 50\% of the recipe steps (enough context) are observed, the model's performance starts to improve.

\subsection{Amount of Training Data}\label{sec:knowledgeTransfer}

\begin{table}[!htb]
\centering
\resizebox{\columnwidth}{!}{
\setlength{\tabcolsep}{6.3pt}
\begin{tabular}{@{}lrrrrr@{}}
\toprule
{Method} & {ING} & {VERBS} & {BLEU1} & {BLEU4} & {METEOR} \\ \midrule
ours text (100\%)&\cellcolor[HTML]{FF9CBD}26.09 &\cellcolor[HTML]{FF9CBD}27.19 &\cellcolor[HTML]{FF9CBD}26.78 &\cellcolor[HTML]{FF9CBD}3.30 &\cellcolor[HTML]{FF9CBD}17.97 \\ 
ours text (50\%) &\cellcolor[HTML]{FFACC7}23.01 &\cellcolor[HTML]{FFADC9}24.90 &\cellcolor[HTML]{FFA9C6}25.05 &\cellcolor[HTML]{FFB9D1}2.42 &\cellcolor[HTML]{FFA4C2}16.98 \\ 
ours text (25\%) &\cellcolor[HTML]{FFBED3}19.43 &\cellcolor[HTML]{FFB5CE}23.83 &\cellcolor[HTML]{FFB5CE}23.54 &\cellcolor[HTML]{FFC6D9}2.03 &\cellcolor[HTML]{FFACC8}16.05 \\ 
ours text (0\%) &\cellcolor[HTML]{FFF5F8}5.80 &\cellcolor[HTML]{FFF5F8}9.42 &\cellcolor[HTML]{FFF5F8}10.58 &\cellcolor[HTML]{FFF5F8}0.24 &\cellcolor[HTML]{FFF5F8}6.80 \\ 
\bottomrule
\end{tabular}} 
\caption{
Evaluations on Tasty V1's test set for the textual model when the number of training recipes varies. Performance drops when the amount of pre-training decreases.
} 
\label{fig:tasty_knowledge}
\end{table}
 
At the core of our method is the transfer of knowledge from text resources to solve a challenging visual problem. We evaluate the effectiveness of the knowledge transfer by varying the amount of training data from Recipe1M to be used for pre-training. We train our method with 100\%, 50\%, and 25\% of the Recipe1M training set. We present our results in Table~\ref{fig:tasty_knowledge}. Looking at the averaged scores over all the predicted steps on the Tasty Videos dataset, we observe a decrease in all evaluation measures as we limit the amount of data from Recipe1M (see \emph{`ours text'} 100\%, 50\%, 25\%, and 0\%), with the most significant decrease occurring for the BLEU4 score. When using less text data, our method's performance decreases from the 3.30 of the BLEU4 score to 2.42 when half of the Recipe1M is used and to 2.03 when a quarter of the dataset is used. While we observe a similar decrease in the ingredient detection scores, the decrease in BLEU1, METEOR, and verb scores remains less significant. If there is no pre-training, \ie, when the model is learned only on text from Tasty Videos V1 (\emph{`ours text (0\%)'}), the decrease in scores is noticeable for all evaluation criteria. These results verify that pre-training has a significant effect on our method's performance.

\begin{table}[t] 
\resizebox{\columnwidth}{!}{
\setlength{\tabcolsep}{3.9pt} 
\begin{tabular}{@{}lrr|rr|rr|rr|rr@{}}
\toprule 
& \multicolumn{2}{c}{ING} & \multicolumn{2}{c}{VERB} & \multicolumn{2}{c}{BLEU1} & \multicolumn{2}{c}{BLEU4} & \multicolumn{2}{c}{METEOR} \\ \midrule
\multicolumn{1}{l}{step:} & Ld & +Lr & Ld & +Lr & Ld & +Lr & Ld & +Lr & Ld & +Lr \\ \midrule 
\multicolumn{1}{l}{next}&\cellcolor[HTML]{FFBAD1}32.3 &\cellcolor[HTML]{FFBAD1}33.5 &\cellcolor[HTML]{FFBAD1}26.1 &\cellcolor[HTML]{FFBAD1}26.7 &\cellcolor[HTML]{FFBAD1}22.1 &\cellcolor[HTML]{FFBAD1}22.8 &\cellcolor[HTML]{FFBAD1}4.3 &\cellcolor[HTML]{FFBAD1}4.4 &\cellcolor[HTML]{FFBAD1}13.4 &\cellcolor[HTML]{FFBAD1}13.7 \\ 
\multicolumn{1}{l}{next+1} &\cellcolor[HTML]{FFDAE7}24.6 &\cellcolor[HTML]{FFD5E3}28.4 &\cellcolor[HTML]{FFE4ED}19.1 &\cellcolor[HTML]{FFE5ED}16.0 &\cellcolor[HTML]{FFE5ED}16.1 &\cellcolor[HTML]{FFE9F0}14.0 &\cellcolor[HTML]{FFE8EF}2.2 &\cellcolor[HTML]{FFF0F5}1.0 &\cellcolor[HTML]{FFE6EF}9.2 &\cellcolor[HTML]{FFE9F0}7.6 \\ 
\multicolumn{1}{l}{next+2} &\cellcolor[HTML]{FFEEF4}19.6 &\cellcolor[HTML]{FFE8F0}24.5 &\cellcolor[HTML]{FFF2F6}16.6 &\cellcolor[HTML]{FFF1F6}12.5 &\cellcolor[HTML]{FFF3F7}13.9 &\cellcolor[HTML]{FFF3F7}11.8 &\cellcolor[HTML]{FFF5F8}1.5 &\cellcolor[HTML]{FFF6F9}0.6 &\cellcolor[HTML]{FFF3F7}7.8 &\cellcolor[HTML]{FFF3F7}6.1 \\ 
\multicolumn{1}{l}{next+3} &\cellcolor[HTML]{FFF8FA}16.5 &\cellcolor[HTML]{FFF8FA}20.9 &\cellcolor[HTML]{FFF8FA}15.2 &\cellcolor[HTML]{FFF8FA}10.3 &\cellcolor[HTML]{FFF8FA}12.9 &\cellcolor[HTML]{FFF8FA}10.5 &\cellcolor[HTML]{FFF8FA}1.3 &\cellcolor[HTML]{FFF8FA}0.4 &\cellcolor[HTML]{FFF8FA}7.2 &\cellcolor[HTML]{FFF8FA}5.2 \\ 
\bottomrule
\end{tabular}}
\caption{
Comparisons for the decoder, $L_{d}$ and decoder + recipe loss, $+L_{r}$, on the Recipe1M's test set. 
} 
\label{tab:modalComp} 
\end{table}

\begin{table}[t] 
\resizebox{\columnwidth}{!}{
\setlength{\tabcolsep}{3.9pt} 
\begin{tabular}{@{}lrr|rr|rr|rr|rr@{}}
\toprule 
& \multicolumn{2}{c}{ING} & \multicolumn{2}{c}{VERB} & \multicolumn{2}{c}{BLEU1} & \multicolumn{2}{c}{BLEU4} & \multicolumn{2}{c}{METEOR} \\ \midrule
\multicolumn{1}{l}{step:} & gr & b & gr & b & gr & b & gr & b & gr & b \\ \midrule 
\multicolumn{1}{l}{next} &\cellcolor[HTML]{FFBAD1}32.3 &\cellcolor[HTML]{FFBAD1}33.1 &\cellcolor[HTML]{FFBAD1}26.1 &\cellcolor[HTML]{FFBAD1}26.9 &\cellcolor[HTML]{FFBAD1}22.1 &\cellcolor[HTML]{FFBAD1}22.8 &\cellcolor[HTML]{FFBAD1}4.3 &\cellcolor[HTML]{FFBAD1}4.6 &\cellcolor[HTML]{FFBAD1}13.4 &\cellcolor[HTML]{FFBAD1}13.9 \\ 
\multicolumn{1}{l}{next+1} &\cellcolor[HTML]{FFDAE7}24.6 &\cellcolor[HTML]{FFD9E6}26.2 &\cellcolor[HTML]{FFE4ED}19.1 &\cellcolor[HTML]{FFE3ED}20.2 &\cellcolor[HTML]{FFE5ED}16.1 &\cellcolor[HTML]{FFE4ED}17.2 &\cellcolor[HTML]{FFE8EF}2.2 &\cellcolor[HTML]{FFE8F0}2.4 &\cellcolor[HTML]{FFE6EF}9.2 &\cellcolor[HTML]{FFE6EE}9.9 \\ 
\multicolumn{1}{l}{next+2} &\cellcolor[HTML]{FFEEF4}19.6 &\cellcolor[HTML]{FFEDF3}21.3 &\cellcolor[HTML]{FFF2F6}16.6 &\cellcolor[HTML]{FFF1F6}17.7 &\cellcolor[HTML]{FFF3F7}13.9 &\cellcolor[HTML]{FFF2F6}15.2 &\cellcolor[HTML]{FFF5F8}1.5 &\cellcolor[HTML]{FFF3F7}1.8 &\cellcolor[HTML]{FFF3F7}7.8 &\cellcolor[HTML]{FFF2F7}8.6 \\ 
\multicolumn{1}{l}{next+3} &\cellcolor[HTML]{FFF8FA}16.5 &\cellcolor[HTML]{FFF8FA}18.2 &\cellcolor[HTML]{FFF8FA}15.2 &\cellcolor[HTML]{FFF8FA}16.3 &\cellcolor[HTML]{FFF8FA}12.9 &\cellcolor[HTML]{FFF8FA}14.1 &\cellcolor[HTML]{FFF8FA}1.3 &\cellcolor[HTML]{FFF8FA}1.5 &\cellcolor[HTML]{FFF8FA}7.2 &\cellcolor[HTML]{FFF8FA}7.9 \\ 
\bottomrule
\end{tabular}}
\caption{
We compare greedy (gr) and beam (b) search when decoding sentences on the Recipe1M's test set. 
} 
\label{tab:beamSearch} 
\end{table}

\begin{figure*}[!htb] 
\centering 
\includegraphics[width=0.99\textwidth]{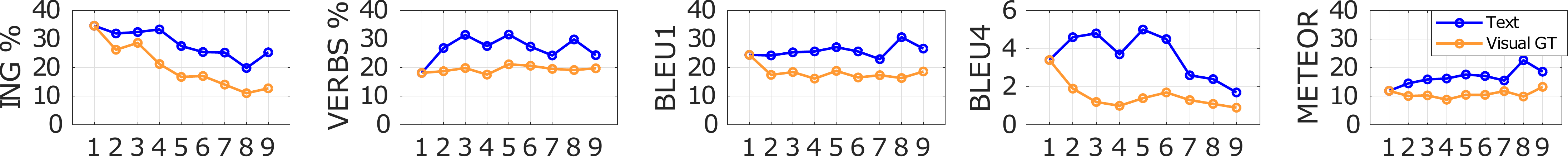}
\caption{
We compare our visual model when tested with GT segments, \emph{`Visual GT'}, and our textual model, \emph{`Text'}, on the Tasty Videos V1's test set for next-step predictions using the recall of predicted ingredients and verbs as well as sentence scores. Compared to our text-based model, our visual model has lower performance but follows similar trends.
} 
\label{fig:tasty_all_res} 
\end{figure*} 
 
\subsection{Loss Variants}

We train our recipe network using a decoder loss, $L_{d}$ (Eq. \ref{eq:objective_decoder}), and additionally propose a recipe loss, $L_{r}$ (Eq. \ref{eq:objective_recipe}), in Section~\ref{sec:model}. Table~\ref{tab:modalComp} presents our experiments analyzing the influence of the recipe loss evaluated on the test set of Recipe1M. Overall, the recipe loss $L_{r}$, as expected, encourages and increases the next-step prediction performance. The increase is more than 0.5\% for verb and ingredient recall and 0.7, 0.1, and 0.3 for the BLEU1, BLEU4, and METEOR scores. However, using the recipe loss significantly decreases the scores in the further stages, particularly for sentence scores. The most significant decrease is observed for the BLEU4 scores, which decrease to 0.4 for four steps into the future, \emph{`next+3'}. Only the ingredient recall does not significantly decrease and is at least 4\% higher than our model trained with the decoder loss alone. However, upon closer inspection of individual predictions, we see that the recipe loss has a tendency to promote repetitive outputs. Repetitions are a common but undesirable outcome for natural language generation~\cite{holtzman2019curious}. In this case, as consecutive steps are more likely to describe the handling of common items or ingredients, the ingredient recall is not as directly affected. 
 
Beam search~\cite{1977cmurep} is known to improve the performance of text-based generation algorithms~\cite{freitag2017beam}. In Table~\ref{tab:beamSearch}, we evaluate our model, trained with decoder loss, $L_{d}$, using greedy and beam search. Greedy decoding selects the words with the highest probability at every decoding step. Beam search keeps track of a beam of $k$ possible generations and updates them at every step of decoding by ranking them according to the model likelihood. Although it is $k$ times more expensive than greedy search, it improves the performance. In our experiments we use $k=5$. This improves both ingredient and verb scores by 0.8\%, and our sentence scores by 0.7, 0.3, and 0.5 for BLEU1, BLEU4, and METEOR, respectively. We observe similar improvements over the later steps as well. We report the beam search encoding results only in Table \ref{tab:beamSearch}, while in our other results tables, out of fairness, we use greedy search.

\subsection{State-of-the-Art Comparisons}
 
 \begin{figure*}[t] 
\centering 
\includegraphics[width=0.41\columnwidth]{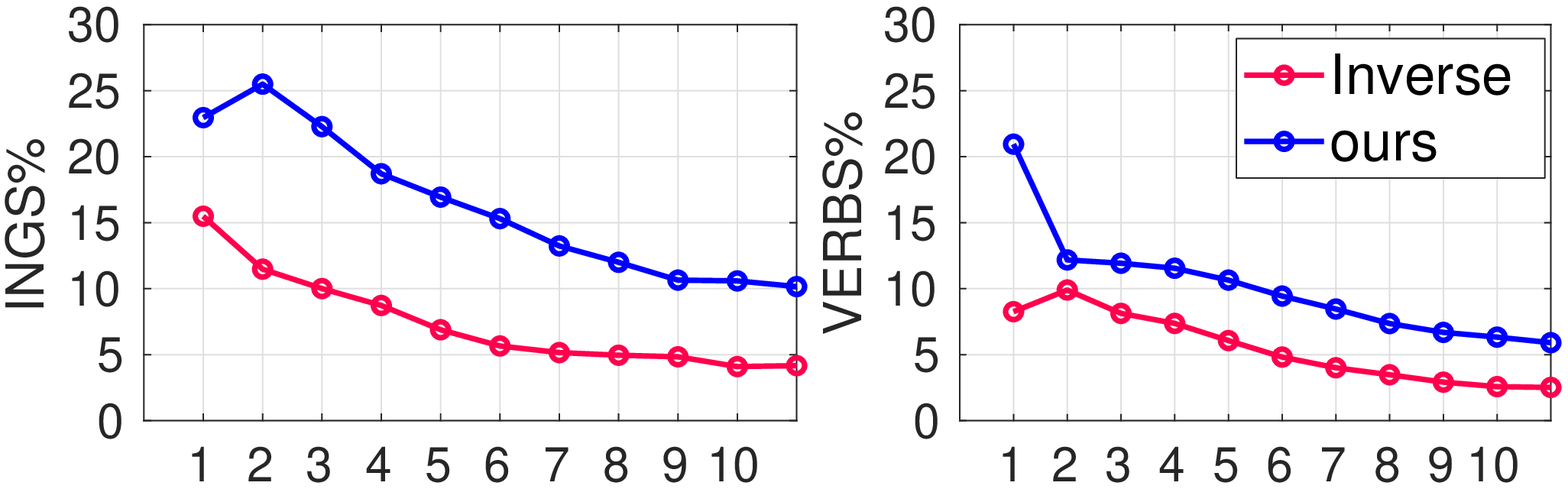}
\includegraphics[width=0.58\columnwidth]{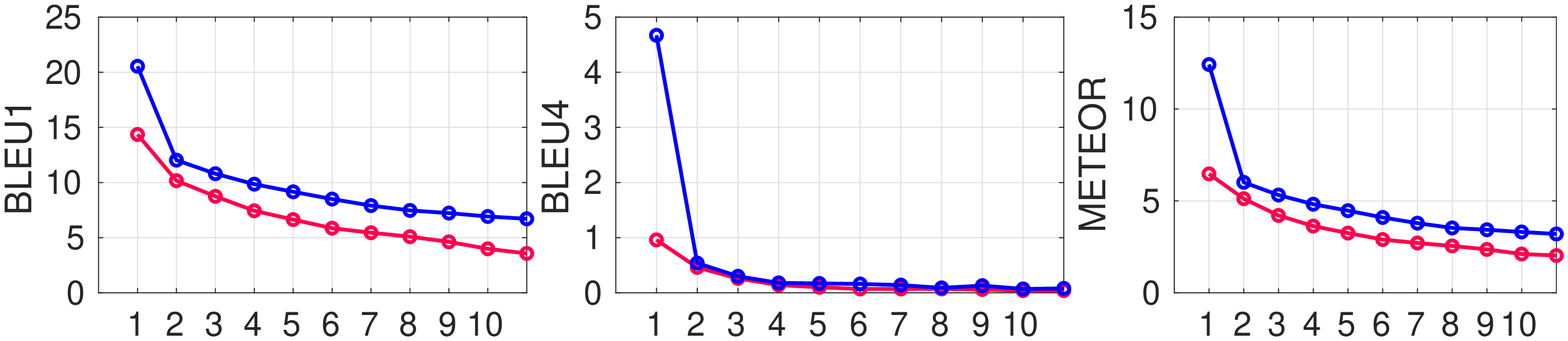}
\caption{
We evaluate our model for recipe generation given ingredients and compare our performance to a recipe generation network, `Inverse'~\cite{salvador2019inverse}, on a subset of Recipe1M.} 
\label{fig:inverseFull} 
\end{figure*} 

\begin{figure*}[t]
\centering 
\includegraphics[width=0.41 \columnwidth]{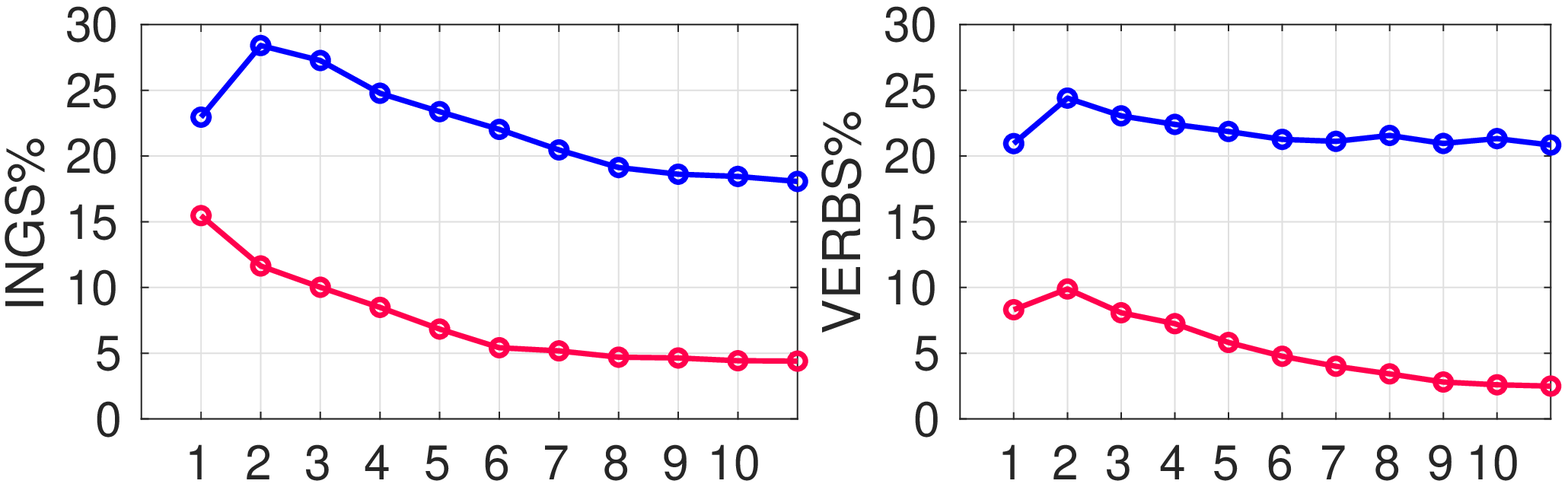}
\includegraphics[width=0.58\columnwidth]{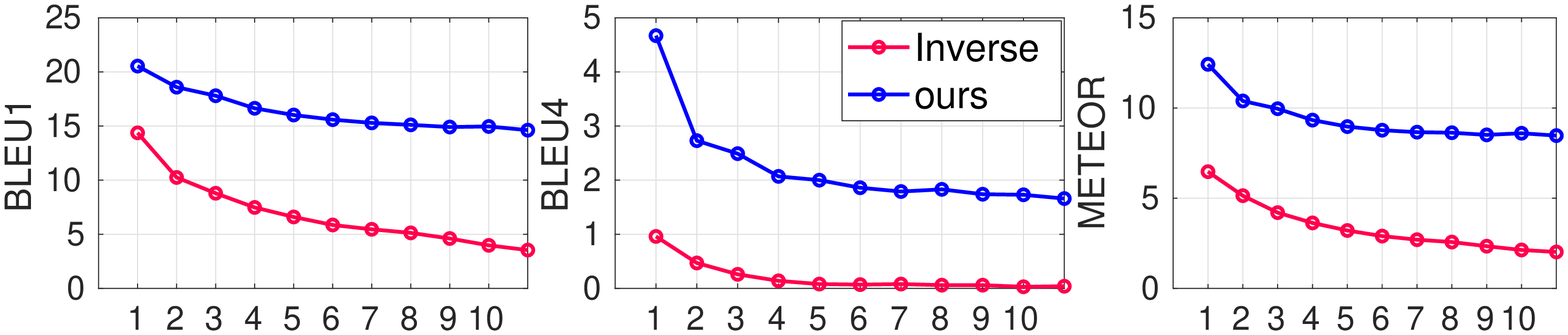}
\caption{
We employ the recipe generation network `Inverse'~\cite{salvador2019inverse} for next-step prediction and compare our performance. Both methods are tested on a subset of Recipe1M. 
}
\label{fig:inverseNextstep} 
\end{figure*} 

There are several works~\cite{kiddon2016globally,salvador2019inverse} that generate instructional text for the cooking domain. Similar to our works, such approaches also target learning procedural knowledge in text. Inverse cooking~\cite{salvador2019inverse} generates recipes for food images. It uses a two-stage transformer approach: the first transformer predicts ingredients from a recipe picture while the second transformer generates the recipe from these ingredients. Inverse cooking uses a subset of Recipe1M's dataset by removing the recipes that contain no images and those with fewer than two ingredients or steps. For a fair comparison, we train and test our model using this subset, which features 252.5k and 54.5k recipes for training and testing, respectively. Their network is limited to generating a maximum of 150 words per recipe. Since our model does not have limitations regarding the generations, for fair comparison, we also truncate our model's generations after 150 words. We compare our model to the publicly released Inverse cooking model by directly providing the ground truth ingredients into their recipe generation transformer.
 
We first evaluate our model's capabilities in recipe generation and compare our generations to Inverse cooking in Figure~\ref{fig:inverseFull} for individual steps. Both methods are evaluated using GT ingredients. Overall, our model outperforms \emph{`Inverse'} for recipe generation for individual steps. Our method's performance is significantly higher for ingredient recall by at least 10\% for all steps. Similarly, for the BLEU4 score, our model's first step prediction is 4.8, while Inverse has only 1. However, for the later steps, we perform comparably in this metric. We observe similar gaps between the first step predictions in the BLEU1, METEOR, and verb scores. 

Next, we evaluate \emph{`Inverse'} for next-step prediction. For each step, we feed the ground truth sentences in the previous steps into Inverse's transformer-based sentence decoder, which generates a sentence for the next step. The results in Figure~\ref{fig:inverseNextstep} show that although the transformer-decoder has access to the complete history of the GT recipe until the next step, our method significantly outperforms this recipe generation network on all scores. For ingredients and verbs, the difference is more than 15\% for all steps. Inverse's performance degrades towards the end of the recipe, while ours stays consistently high for all metrics. The most significant decrease for Inverse is in BLEU4, which decreases to almost zero after the 5th step, whereas ours is invariably above 1.5.

\section{Experiments: Video}

\subsection{Video Predictions on Tasty V1}\label{sec:tastyV1}

\begin{figure} 
\centering 
\includegraphics[ width=0.95\textwidth]{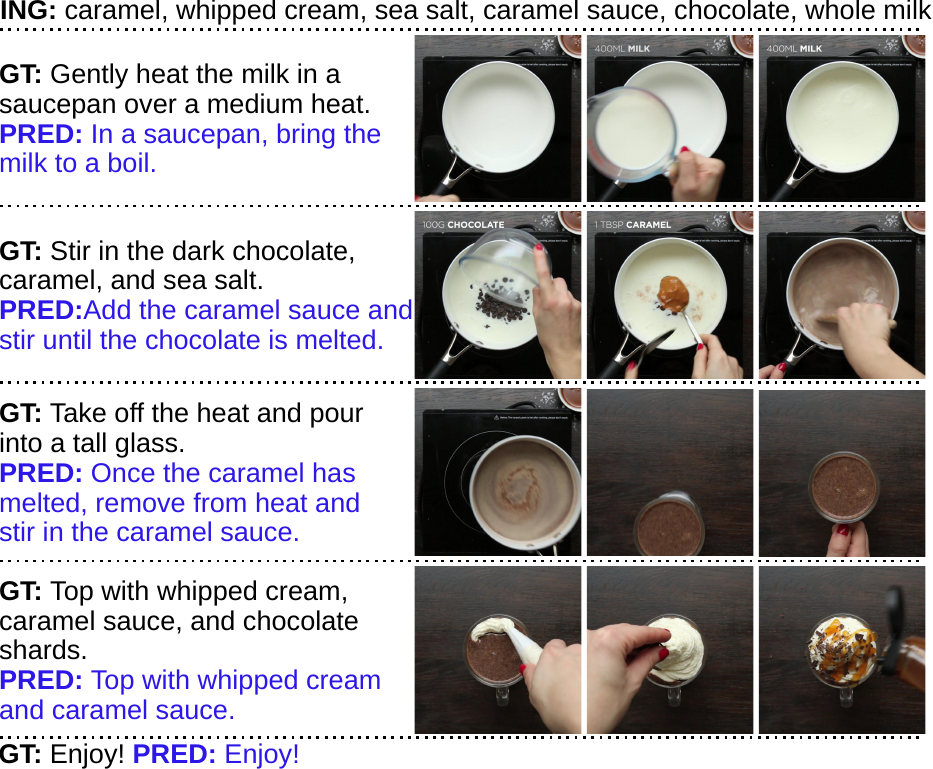} 
\caption{
Next-step predictions from our visual model for \emph{`Salted Caramel Hot Chocolate'} in blue. Note that our model predicts the next steps without having observed the corresponding video segment.} 
\label{fig:overview_example} 
\end{figure}

\begin{figure} [!htb]
\centering 
\includegraphics[ width=0.96\textwidth]{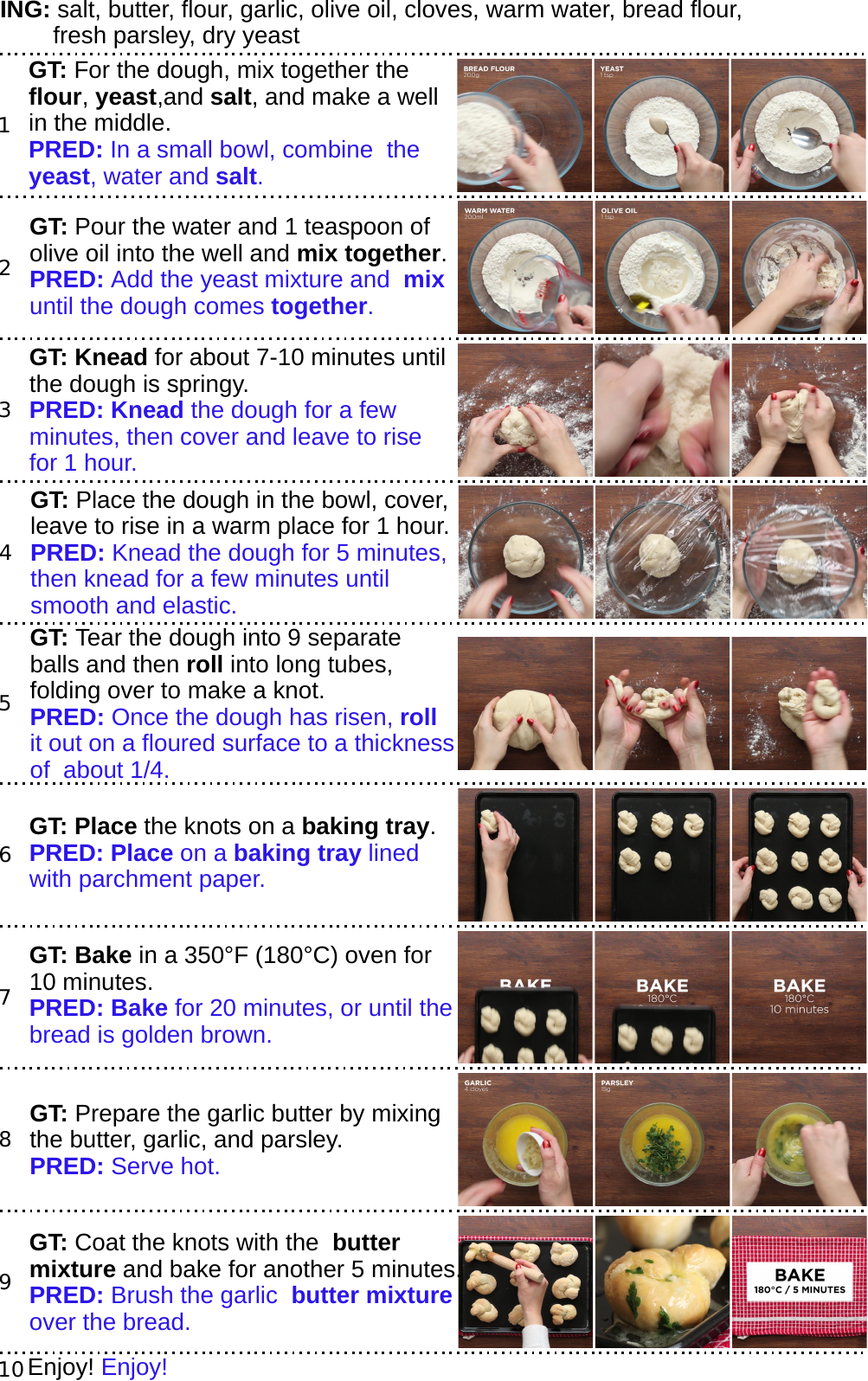} 
\caption{
Next-step predictions for `Garlic Knots' shown in blue. After baking in step 7, our visual model predicts that the dish should be served. Yet when presented with visual evidence of the garlic butter in step 8, it correctly predicts that the knots should be brushed with the mixture in step 9.
} 
\label{fig:overview_example2} 
\end{figure}

We first evaluate our model for making predictions on video inputs on Tasty Videos V1's test set. To explore the importance of video partitioning, we consider the following settings for inference, one according to ground truth segments, \emph{`ours visual (GT)'}, and one based on fixed temporal windows, \emph{`ours visual (window)'}. For temporal window experiments, the videos are partitioned into chunks of fixed-sized windows until the latest observation. We sequentially feed the representations of these chunks obtained from the video encoder into our recipe RNN. Overall, our method is relatively robust to window size, as shown in Table~\ref{fig:windsSel} for BLEU4 scores. We empirically select a window of 170 for Tasty Videos. In both settings, every fifth frame in the GT or window-based segments is sampled, and their visual features are fed into the video encoder. The representations from the video encoder are then fed to the recipe RNN as context vectors. Through the video encoder, our model can interpret visual evidence, and make plausible predictions of the next steps; these are exemplified in Figures~\ref{fig:overview_example} and \ref{fig:overview_example2}, where it can be seen that our visual model corrects itself after observing new evidence.

\begin{table}[t] 
\centering
\resizebox{\columnwidth}{!}{
\setlength{\tabcolsep}{3.5pt}
\begin{tabular}{@{}lrrrrr@{}}
\toprule
Method & ING & VERBS & BLEU1 & BLEU4 & METEOR \\ \midrule
S2VT~\cite{venugopalan2015sequence} (GT) &\cellcolor[HTML]{FFE0EB}7.59 &\cellcolor[HTML]{FFD0E0}19.18 &\cellcolor[HTML]{FFCFDF}18.03 &\cellcolor[HTML]{FFDDE9}1.10 &\cellcolor[HTML]{FFD9E6}9.12 \\ 
S2VT~\cite{venugopalan2015sequence}, next (GT) &\cellcolor[HTML]{FFF4F7}1.54 &\cellcolor[HTML]{FFF4F7}10.66 &\cellcolor[HTML]{FFF4F7}9.14 &\cellcolor[HTML]{FFF4F7}0.26 &\cellcolor[HTML]{FFF3F7}5.59 \\ 
End-to-end~\cite{zhou2018end}& - &- &- &\cellcolor[HTML]{FFEEF3}0.54 &\cellcolor[HTML]{FFF4F7}5.48 \\ \midrule
ours visual (GT) &\cellcolor[HTML]{FFA8C5}20.40 &\cellcolor[HTML]{FFD0E0}19.18 &\cellcolor[HTML]{FFC8DA}19.05 &\cellcolor[HTML]{FFD1E0}1.48 &\cellcolor[HTML]{FFC3D7}11.78 \\
ours visual (window) &\cellcolor[HTML]{FFBAD1}16.66 &\cellcolor[HTML]{FFDDE8}17.08 &\cellcolor[HTML]{FFD1E1}17.59 &\cellcolor[HTML]{FFD9E6}1.23 &\cellcolor[HTML]{FFC9DB}11.00 \\ \midrule
ours text &\cellcolor[HTML]{FF8CB3}26.09 &\cellcolor[HTML]{FF8CB3}27.19 &\cellcolor[HTML]{FF8CB3}26.78 &\cellcolor[HTML]{FF8CB3}3.30 &\cellcolor[HTML]{FF8CB3}17.97 \\
ours text noING &\cellcolor[HTML]{FFDBE7}9.04 &\cellcolor[HTML]{FFBBD2}22.00 &\cellcolor[HTML]{FFC1D6}20.11 &\cellcolor[HTML]{FFE3EC}0.92 &\cellcolor[HTML]{FFB8D0}13.07 \\
ours video-text &\cellcolor[HTML]{FF9FBF}22.27 &\cellcolor[HTML]{FFB0CA}23.35 &\cellcolor[HTML]{FFB5CE}21.75 &\cellcolor[HTML]{FFB2CC}2.33 &\cellcolor[HTML]{FFAFCA}14.09 \\
\bottomrule
\end{tabular}} 
\caption{
Evaluations on Tasty Videos V1 for our visual and text-based model along with comparisons against video captioning methods~\cite{venugopalan2015sequence,zhou2018end}. Performance drops when fixed-sized window-based segments, \emph{`ours visual (window)'}, are used compared to using GT segments, \emph{`ours visual (GT)'}.
Ingredient inputs are important for our model's success. Our method performs better than video captioning.
} 
\label{fig:tasty_res}
\end{table}

\begin{table}[t] 
\centering
\resizebox{\columnwidth}{!}{
\setlength{\tabcolsep}{2.6pt}
\begin{tabular}{@{}llllllllllll@{}}
\toprule
window s. & 30 & 50 & \textbf{70} & 90 & 110 & 130 & 150 & \textbf{170} & 190 & 210 & 230 \\ \midrule 
Tasty V1 & 0.75 & 0.90 & 0.93 & 1.06 & 1.18 & 1.09 & 1.07 & \textbf{1.23} & 1.09 & 1.19 & 1.06 \\
YouCookII & 0.60 & 1.10 & \textbf{1.38} & 1.32 & 1.32 & 1.28 & 1.30 & 1.20 & 1.17 & 1.20 & 1.22 \\ \bottomrule 
\end{tabular}} 
\caption{Window size selection on the Tasty V1 and YouCookII datasets. Reported are the BLEU4 scores. } 
\label{fig:windsSel}
\end{table}

The results are shown in Table~\ref{fig:tasty_res}. Compared to using ground truth segments, \emph{`ours visual (GT)'}, using fixed window segments, \emph{`ours visual (window)'}, results in a decrease in performance, with the most extreme drop on the most challenging sentence score, BLEU4 (around 17\%), and ingredient scores (around 18\%). For the verb, BLEU1, and METEOR scores, the decrease is not as big (lower than 10\%).

In Table~\ref{fig:tasty_res}, our text-based results are presented as upper-bound, \emph{`ours text'}. Given that our model is first trained on text and then transferred to video, the drop in performance from text to video is as expected. The video results, however, follow similar trends as the text; see, for example, Figure~\ref{fig:tasty_all_res}, where we provide step-wise comparisons of our textual and visual models (GT). We further investigate the influence of the ingredients on the performance of our method. When ingredients are not provided, \emph{`ours text noING'}, our method fails to make plausible predictions. The performance decrease is mainly noticeable in the ingredient scores and the BLEU4 scores.
 
In some instructional scenarios, there may be semi-aligned text that accompanies the video, \eg, narrations. We test such a setting by training the sentence and video encoder, as well as sentence decoder and recipe RNN jointly, to make future step predictions. For this, the sentence and video context vectors are first concatenated and then passed through a linear layer before feeding them as input to the Recipe RNN. Overall, the results are better than our video-alone results but not better than our text-alone results (see \emph{`ours video-text'} in Table~\ref{fig:tasty_res}). Even with joint training, it is still challenging to make improvements, which we attribute to the diversity in our videos and the variations in the text descriptions for similar visual inputs. On the other hand, when there is accompanying text, our model can be adapted easily and improve prediction performance.

\subsection{Video Predictions on YouCookII}\label{sec:youcook2res}

\begin{table} 
\centering
\resizebox{\columnwidth}{!}{
\setlength{\tabcolsep}{3.6pt}
\begin{tabular}{@{}lrrrrr@{}}
\toprule
Method & ING & VERBS & BLEU1 & BLEU4 & METEOR \\ \midrule
End-to-end (GT)~\cite{zhou2018end} &- &- &- &\cellcolor[HTML]{FFDBE7}0.87 &\cellcolor[HTML]{FFCDDE}8.15 \\ 
TempoAttn (GT)~\cite{yao2015describing} &- &- &- &\cellcolor[HTML]{FFC4D8}1.42 &\cellcolor[HTML]{FFA7C5}11.20 \\ 
ours visual (GT) &\cellcolor[HTML]{FFC3D7}21.36 &\cellcolor[HTML]{FFC9DB}27.55 &\cellcolor[HTML]{FFE3EC}23.71 &\cellcolor[HTML]{FFBAD1}1.66 &\cellcolor[HTML]{FFA3C2}11.54 \\ \midrule
End-to-end~\cite{zhou2018end} &- &- &- &\cellcolor[HTML]{FFF4F7}0.08 &\cellcolor[HTML]{FFF4F7}4.62 \\ 
TempoAttn~\cite{yao2015describing} &- &-&- &\cellcolor[HTML]{FFEEF4}0.30 &\cellcolor[HTML]{FFDFEA}6.58 \\ 
ours visual (window) &\cellcolor[HTML]{FFF4F7}17.64 &\cellcolor[HTML]{FFF4F7}25.11 &\cellcolor[HTML]{FFF4F7}22.55 &\cellcolor[HTML]{FFC6D9}1.38 &\cellcolor[HTML]{FFAEC9}10.71 \\ \midrule
ours text &\cellcolor[HTML]{FF8CB3}24.60 &\cellcolor[HTML]{FF8CB3}29.39 &\cellcolor[HTML]{FF8CB3}26.49 &\cellcolor[HTML]{FF8CB3}2.66 &\cellcolor[HTML]{FF8CB3}13.31 \\ 
\bottomrule
\end{tabular}} 
\caption{
Evaluation of our visual and text-based models and comparison against two video captioning methods~\cite{yao2015describing,zhou2018end} on YouCookII's validation set. Note that in this comparison, we are anticipating the NLP description of the next step, while the captioning methods are applied directly to future (unseen to us) video footage. Even though both methods, including~\cite{zhou2018end}, which is state-of-the-art for dense captioning, have access to the video and we do not, we still perform better in both the BLEU4 and METEOR scores.} 
\label{fig:YoucookII_res_supervised}
\end{table}

To further validate the effectiveness of our model on publicly available datasets with sentence-level annotations, we also evaluate it on the YouCookII dataset. Table~\ref{fig:YoucookII_res_supervised} compares our visual model evaluated with ground truth segments, \emph{`ours visual (GT)'}, and temporal windows, \emph{`ours visual (window)'}, on YouCookII's validation set. For the \emph{`ours visual (window)'} experiments, a window of 70 frames is empirically selected, see Table~\ref{fig:windsSel}. YouCookII includes a large number of background frames that are irrelevant to the recipe steps. We suspect using large windows misses important cues for the YouCookII steps. 

Similar to our observations on Tasty V1, using window segments, \emph{`ours visual (window)'}, instead of ground truth, \emph{`ours visual (GT)'}, results in decreased accuracy. The largest decrease is observed for the BLEU4 and ingredient scores, by 16\% and 17\%, respectively, highlighting the importance of achieving high scores for these metrics. The decrease for BLEU1 is the smallest, by 4\%. Comparing the performance of our visual with the textual model, \emph{`ours text'}, the textual results are better overall on Tasty V1 than YouCookII.

\subsection{Video Predictions on Tasty V2}\label{sec:tastyV2}

\begin{table} 
\centering
\resizebox{\columnwidth}{!}{
\setlength{\tabcolsep}{5.4pt}
\begin{tabular}{@{}lrrrrr@{}}
\toprule
 & ING & VERB & BLEU1 & BLEU4 & METEOR \\ \midrule 
video (GT) &\cellcolor[HTML]{FFF3F7}19.33 &\cellcolor[HTML]{FFF6F9}16.86 &\cellcolor[HTML]{FFF5F8}19.34 &\cellcolor[HTML]{FFF2F7}1.71 &\cellcolor[HTML]{FFEEF4}12.64 \\
video (window) &\cellcolor[HTML]{FFF7F9}18.60 &\cellcolor[HTML]{FFF7F9}16.47 &\cellcolor[HTML]{FFF7F9}18.60 &\cellcolor[HTML]{FFF7F9}1.42 &\cellcolor[HTML]{FFF7F9}11.75 \\
video (proposal) &\cellcolor[HTML]{FFF4F8}19.12 &\cellcolor[HTML]{FFF7F9}16.37 &\cellcolor[HTML]{FFF7F9}18.60 &\cellcolor[HTML]{FFF4F7}1.63 &\cellcolor[HTML]{FFF6F9}11.84 \\ \midrule 
text &\cellcolor[HTML]{FFABC7}28.93 &\cellcolor[HTML]{FFABC7}27.37 &\cellcolor[HTML]{FFABC7}26.97 &\cellcolor[HTML]{FFABC7}4.61 &\cellcolor[HTML]{FFABC7}18.26 \\
\bottomrule
\end{tabular}
} 
\caption{
Evaluation of the visual and text-based models and comparisons using different segments on Tasty V2. 
} 
\label{tab:tastyv2}
\end{table}
 
\begin{figure}
\centering 
\includegraphics[width=1\linewidth]{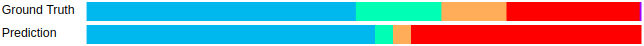} 
\caption{Example proposals from a proposal decoder~\cite{zhou2018end} trained on Tasty V2.}
\label{fig:discrete_proposals} 
\end{figure}

In addition to using fixed window-based segments, we train the transformer-based proposal decoder from Zhou et al.~\cite{zhou2018towards} on the Tasty V2 dataset to generate segment proposals. First, using non-maximum suppression, at each iteration, the longest proposal is selected, and those with an IoU with the longest proposal that is greater than a threshold of 0.2 are discarded. Then, to obtain non-overlapping proposals, the overlapping regions among the overlapping proposals are divided w.r.t.\ their lengths; an example can be seen in Figure~\ref{fig:discrete_proposals}. We evaluate the quality of the proposals using mean intersection over union (IoU), which is 39.59\%.

The results are shown in Table~\ref{tab:tastyv2}. To validate our method's performance on Tasty V2, we use ground truth segments, \emph{`video (GT)'}, fixed temporal windows, \emph{`video (window)'}, and segment proposals, \emph{`video (proposal)'}, as input to our video encoder during inference. Compared to using temporal windows, proposals improve ingredient scores by 2\% and BLEU4 by 12\%. The other scores show slight differences. This small performance gap between the window and proposal-based segments indicates the difficulty of partitioning our videos and motivates exploring more robust ways of generating segments/proposals. The text-based results, \emph{`ours text'}, maintain the highest scores in all metrics, encouraging us to further develop visual models to tackle in anticipation of our challenging zero-shot dataset.

\subsection{Supervised vs. Zero-Shot Learning}

\begin{table}[t]
\centering
\resizebox{\columnwidth}{!}{
\setlength{\tabcolsep}{2.0pt}
\begin{tabular}{@{}lrrrrr@{}}
\toprule
{Method} & {ING} & {VERBS} & {BLEU1} & {BLEU4} & {METEOR} \\ \midrule 
sup. visual (GT)&\cellcolor[HTML]{FFB9D0}20.93 &\cellcolor[HTML]{FFC8DA}24.76 &\cellcolor[HTML]{FFC8DA}22.11 &\cellcolor[HTML]{FFCDDE}1.21 &\cellcolor[HTML]{FFBFD4}10.66 \\ 
sup. visual (window) &\cellcolor[HTML]{FFC1D5}18.90 &\cellcolor[HTML]{FFD8E5}23.15 &\cellcolor[HTML]{FFD1E0}21.09 &\cellcolor[HTML]{FFD4E3}1.03 &\cellcolor[HTML]{FFC4D8}10.22 \\ 
sup. visual, w/o pre-train &\cellcolor[HTML]{FFF5F8}2.69 &\cellcolor[HTML]{FFF2F6}19.43 &\cellcolor[HTML]{FFF6F9}15.05 &\cellcolor[HTML]{FFF3F7}0.15 &\cellcolor[HTML]{FFF0F5}5.89 \\ 
sup. text &\cellcolor[HTML]{FFABC7}24.56 &\cellcolor[HTML]{FFABC7}27.24 &\cellcolor[HTML]{FFABC7}24.94 &\cellcolor[HTML]{FFABC7}1.99 &\cellcolor[HTML]{FFABC7}12.50 \\ \midrule
zero visual (GT) &\cellcolor[HTML]{FFC5D8}17.77 &\cellcolor[HTML]{FFD9E5}23.11 &\cellcolor[HTML]{FFD5E3}20.61 &\cellcolor[HTML]{FFDCE8}0.84 &\cellcolor[HTML]{FFCBDD}9.51 \\ 
zero visual (window) &\cellcolor[HTML]{FFECF2}6.04 &\cellcolor[HTML]{FFD8E5}23.19 &\cellcolor[HTML]{FFD7E5}20.30 &\cellcolor[HTML]{FFDFEA}0.76 &\cellcolor[HTML]{FFCEDE}9.27 \\ 
zero visual, w/o pre-train &\cellcolor[HTML]{FFF7F9}1.58 &\cellcolor[HTML]{FFF7F9}17.83 &\cellcolor[HTML]{FFF7F9}14.54 &\cellcolor[HTML]{FFF7F9}0.01 &\cellcolor[HTML]{FFF7F9}5.03 \\ 
zero text &\cellcolor[HTML]{FFBDD3}19.90 &\cellcolor[HTML]{FFC7DA}24.86 &\cellcolor[HTML]{FFBED4}23.06 &\cellcolor[HTML]{FFC2D6}1.47 &\cellcolor[HTML]{FFBCD2}10.98 \\ 
\bottomrule
\end{tabular} }
\caption{
Comparison of our zero-shot and supervised setting on YouCookII, computed using 4-fold cross-validation. The supervised results are better overall. Without pre-training on Recipe1M, the performance drop is significant. 
}
\label{fig:YoucookII_res_zero_shot}
\end{table}

YouCookII is a suitable dataset to compare the differences between supervised and zero-shot learning. As the provided splits for this dataset are not zero-shot and overlap in the dishes for training and test, we create our own splits based on distinct dishes for this ablation study. We divide the dataset into four splits based on the 89 dishes, 22 dishes per split, and use three splits for training and half of the videos in the fourth split for testing. In the zero-shot setting, the videos from the other half of the fourth split are unused, while in the supervised setting, they are included as part of the training. 

We report our results as averages over the four cross-folds in Table~\ref{fig:YoucookII_res_zero_shot}. As expected, the predictions are better when the model is trained under a supervised setting than a zero-shot setting. This is true for all inputs, with the same drop as observed previously when moving from text to video inputs and when moving from ground truth video segments to fixed window segments. However, the difference between the supervised versus zero-shot setting (\emph{`sup. visual'} vs.\ \emph{`zero visual'}) is surprisingly much smaller than the difference between a supervised setting with and without pre-training on Recipe1M (\emph{`sup. visual'} vs.\ \emph{`sup. visual w/o pre-train'}). This suggests that having a large corpus for pre-training is more useful than repeated observations for a specific dish.
 
\begin{figure}[t]
\centering 
\includegraphics[width=0.9\columnwidth]{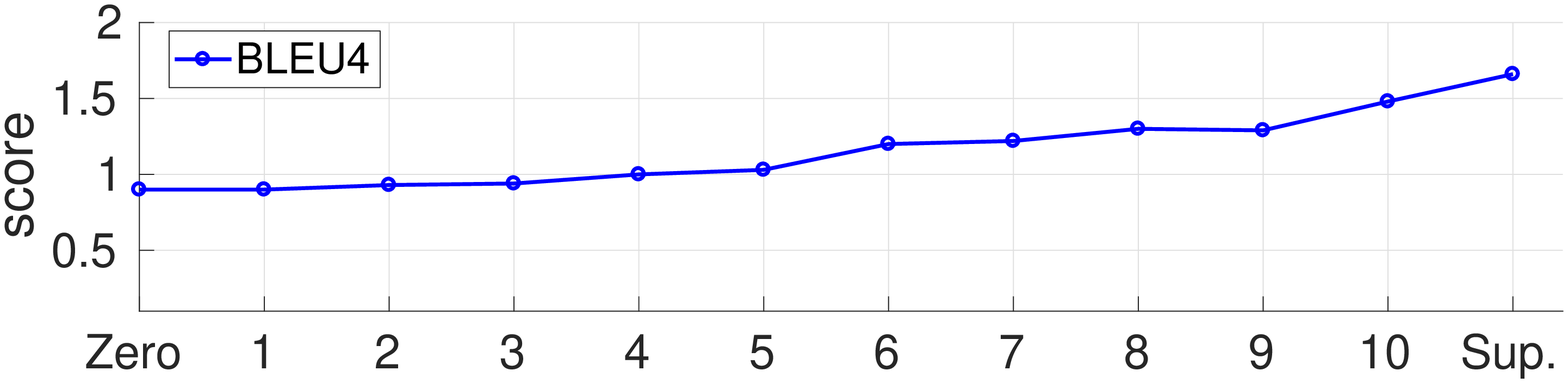} 
\caption{
Zero-shot (\emph{`Zero'}) vs. supervised (\emph{`Sup.'}) comparison on YouCookII when the number of training videos from the same dish is increased. When more videos from the same dish are added into the training set, the BLEU4 score increases.
}
\label{fig:zero_supervised} 
\end{figure}

Figure~\ref{fig:zero_supervised} shows a detailed transition from the zero-shot scenario (no videos about the evaluated dish in the training set) to a one-shot setting (only one similar video) and incremental addition of training videos until the fully supervised case (average of 11 videos from the same dish). One can see that performance increases as more videos are added, indicating that the model is learning and that more than 11 videos (current supervised setting) will further improve the supervised performance.

\subsection{Comparisons to Video Captioning}

We show that knowledge transfer considerably improves our method's predictions, see Sec.~\ref{sec:knowledgeTransfer}. To further validate our claims, we compare our method against different video captioning methods in Tables~\ref{fig:tasty_res}, Table~\ref{fig:tastyv2_segmentation} and ~\ref{fig:YoucookII_res_supervised} for the Tasty Videos V1, V2 and YouCookII datasets, respectively. Unlike predicting future steps, captioning methods generate sentences after observing visual data. In principle, this should be an easier task than predicting the future. 

We compare our model on the validation set of YouCookII against two captioning methods~\cite{yao2015describing,zhou2018end} in Table~\ref{fig:YoucookII_res_supervised}. The End-to-end masked transformer~\cite{zhou2018end} performs dense video captioning by both localizing steps and generating descriptions for these steps. Instead of separating the captioning problem into the two stages of proposal generation and captioning, \cite{zhou2018end} produce proposals and descriptions simultaneously. Their work is composed of a transformer-based~\cite{vaswani2017attention} video encoder for context-aware features, a proposal decoder similar to \cite{zhou2018towards} that localizes action proposal candidates, and finally, a transformer-based decoder that generates captions. TempoAttn~\cite{yao2015describing} is an RNN-based encoder-decoder with attention. A variant of TempoAttn~\cite{yao2015describing} is trained on YouCookII by Zhou~\etal~\cite{zhou2018end} after several changes have been made to the model for a fair comparison, including adding a Bi-LSTM context encoder and adding temporal attention. 

\begin{table}[t] 
\centering
\resizebox{\linewidth}{!}{
\setlength{\tabcolsep}{12.6pt}
\begin{tabular}{@{}lrrr@{}}
\toprule
& BLEU1 & BLEU4 & METEOR \\ \midrule
End-to-end (GT)~\cite{zhou2018end} & 18.33 & 1.14 & 6.73 \\ 
ours visual (GT) & 19.34 & 1.71 & 12.64 \\ \bottomrule
\end{tabular} }
\caption{Captioning results of End-to-end~\cite{zhou2018end} vs. our method evaluated for next-step prediction on the Tasty V2 dataset. Both methods are evaluated using GT segments.
} 
\label{fig:tastyv2_segmentation} 
\end{table}

In Table~\ref{fig:YoucookII_res_supervised}, we see that even though the anticipation task is more difficult than captioning, our method outperforms both of the captioning methods for the BLEU4 and METEOR scores. Compared to the state-of-the-art video captioning method,~\cite{zhou2018end}, our visual model achieves a METEOR score twice as high and a BLEU4 score four times higher. We attribute the better performance of our method compared to the captioning methods to the pre-training on the Recipe1M dataset, which allows our model to generalize. Note that for YouCookII, as we use all the videos in the training set, our training is no longer a zero-shot but a supervised scenario. 

Table~\ref{fig:tasty_res} compares our model against different captioning methods on the Tasty V1 dataset. We also test S2VT~\cite{venugopalan2015sequence}, an RNN-based encoder-decoder, on the ground truth segments of Tasty V1 for captioning. Our visual model outperforms this baseline, especially for ingredient recall, by 13\%, and with an improvement of 0.3 in the BLEU4 score. To highlight the difficulty of predicting future steps compared to captioning, we train S2VT for predicting the next step from the observation of the current step, \emph{`S2VT~\cite{venugopalan2015sequence} next (GT)'}. Our visual model outperforms this variation with a significant margin for all scores. We also test a state-of-the-art video captioning method, \emph{`End-to-end ~\cite{zhou2018end}'}, on Tasty V1 and get a BLEU4 and METEOR score of 0.54 and 5.48, respectively (vs. our future prediction scores 1.23 / 11.00). The poor performance is likely due to the increased dish diversity and the difficulty of our dataset compared to YouCookII.

Finally, we train the End-to-end masked transformer~\cite{zhou2018end} on our Tasty V2 for captioning. Table~\ref{fig:tastyv2_segmentation} compares our model to this work using ground truth segments. Similar to our observations on other datasets, although \emph{`End-to-end~\cite{zhou2018end}}', incorporates context into predictions and predicts the current observation, our method, which predicts the next steps, outperforms it for all sentence scores.

\subsection{Human Ratings}\label{sec:human_study}

\begin{figure}[t]
\centering
 \begin{tabular}{@{}c@{}}
 \includegraphics[width=0.97\textwidth]{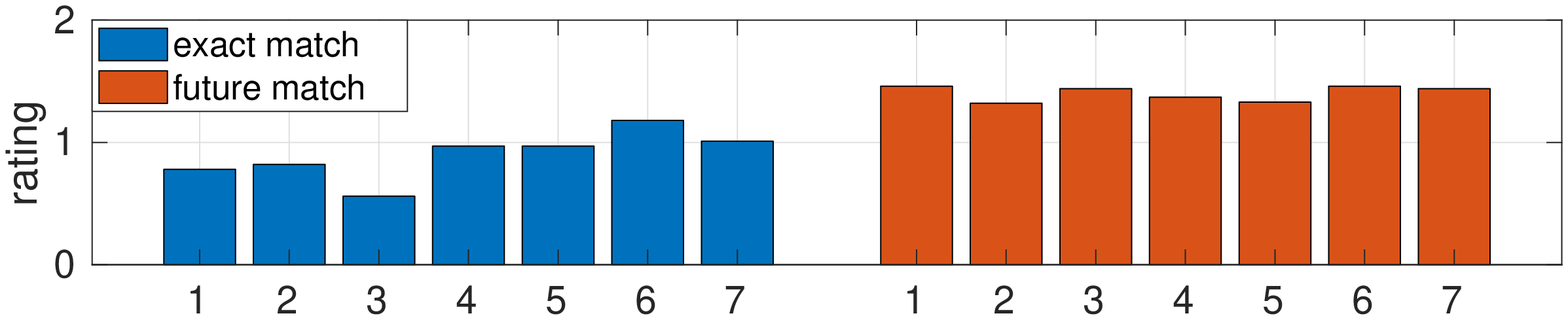}\\
 \small (a) Human ratings.
 \end{tabular}
 \begin{tabular}{@{}c@{}}
 \includegraphics[width=0.97\textwidth]{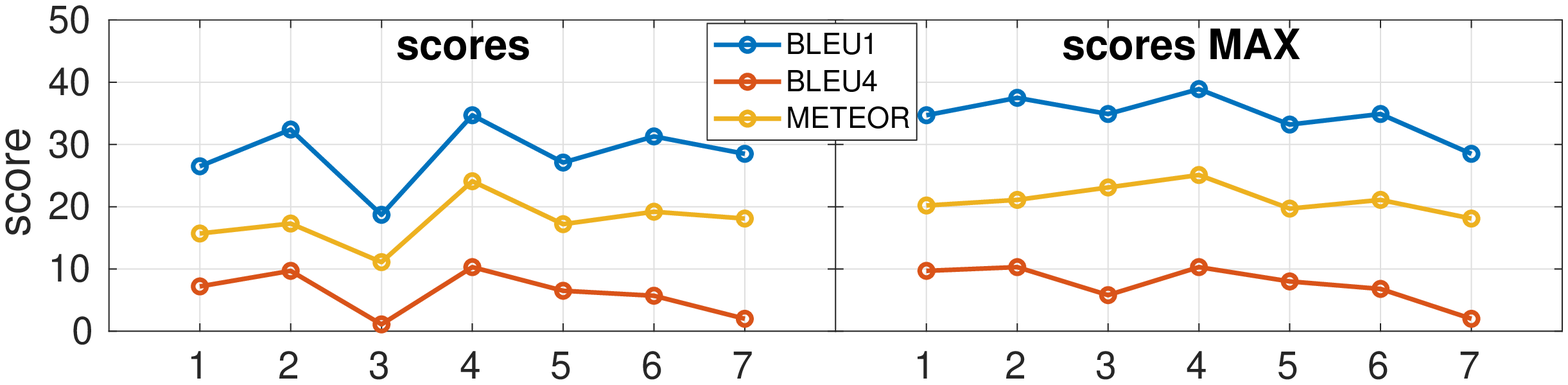}\\
 \small (b) Sentence scores.
 \end{tabular}
\caption{
We conduct a user study and ask human raters to asses how well the predicted sentences match the ground truth sentences. We present the comparison of human ratings (a) versus automated sentence scores (b).
}
\label{fig:human_selected} 
\end{figure}

As automated scores such as BLEU and METEOR are not fully representative of the correctness of the predicted steps, we also ask humans to evaluate our model's predictions. We invite three volunteers to assess how well the anticipated steps match the ground truth with scores 0 (\emph{`not at all'}), 1 (\emph{`somewhat'}), or 2 (\emph{`very well'}). If the prediction receives a score of 0, we additionally ask the participant to judge if the predicted step is still a plausible future prediction, again with the same scores of 0 (\emph{`not at all'}), 1 (\emph{`somewhat'}), or 2 (\emph{`very likely'}). The study is done on a subset of 30 recipes from Recipe1M's test set, each with seven steps. Ratings are compared to the automated sentence scores in Figure~\ref{fig:human_selected}.

In Figure~\ref{fig:human_selected}, the upper graph (a) shows the results for the human raters. In this plot, \emph{`exact match'} corresponds to humans assessing if the predicted steps match the ground truth, \ie the plausibility of the prediction regarding the ground truth sentence. Raters report a score close to 1 for the initial step predictions, indicating that our method, even by only seeing the ingredients, can start predicting plausible steps. Scores increase towards the end of the recipe and are lowest at step 3. \emph{`future match'} corresponds to humans assessing if a step is a plausible future prediction given all previous steps. The average score of the predicted steps being a possible future prediction is consistently high across all steps. Even if the predicted step does not exactly match the ground truth, human raters still consider it possible in the future, including the previously low rating for step 3. Overall, the ratings indicate that the predicted steps are plausible.

The lower graph (b) in Figure~\ref{fig:human_selected} shows automated scores for the same set of recipes used in our user study. The left plot shows the standard scores for the predicted sentences matching the ground truth. Overall, the trends are very similar to the user study, including the low-scoring step 3. To match the second setting of the user study, we compute the sentence scores between the predicted sentence $\hat{s}_j$, and the next four future ground truth steps $\{s_j, s_{j+1}, s_{j+2},s_{j+3}\}$ and select the step with the maximum score as our future match. These scores are plotted in the lower right plot of Figure~\ref{fig:human_selected}. Similar to the second setting in the human study, the sentence scores increase overall.

\section{Conclusion}
In this paper, we posed a new problem setting of zero-shot action anticipation. 
We presented a model that can generalize instructional knowledge from the text domain and be applied to videos. Using this model, we tackled the challenging task of predicting the steps of complex tasks from visual data. Our model produces coherent and plausible future steps from both text and video inputs. Such a task has been to date otherwise not possible because of the scarcity in annotated training data. Our evaluation shows that our anticipation method is more competitive than all other baselines, even when compared against video captioning methods that have access to the visual data. While we score well for keyword recall like ingredients and verbs, our sentence scores, like the challenging BLEU4, are still poor. We believe this highlights the difficulty of our task and thus aim for improvements in our future work. 
 
To complement our new task and model, we presented a diverse dataset of 4022 cooking videos and recipes. All the videos are annotated with the temporal boundaries of the textual recipe steps. Our dataset includes cooking videos with various dish categories, cookware, and ingredients and provides researchers with a rich database to study the challenging zero-shot anticipation problem. We also hope that its diversity will motivate researchers to study tasks beyond anticipation, such as dense video captioning, temporal segmentation, visual grounding, and retrieval. 

Currently, our method only employs textual recipes and ingredient keys. It can be further improved using additional cues such as the amount of ingredients, which are crucial for real-life instructions. However, this will likely require a dedicated architecture for handling such data to keep track of the ingredient amounts. Further improvements could be achieved by aggregating information from multiple similar recipes or from user feedback and comments.

\section*{Acknowledgment}
This research is supported by the National Research Foundation, Singapore under its NRF Fellowship for AI (NRF-NRFFAI1-2019-0001)

\bibliographystyle{ieeetr.bst} 
\bibliography{references}

\begin{IEEEbiography}[{\includegraphics[width=1in,height=1.25in,clip,
keepaspectratio]{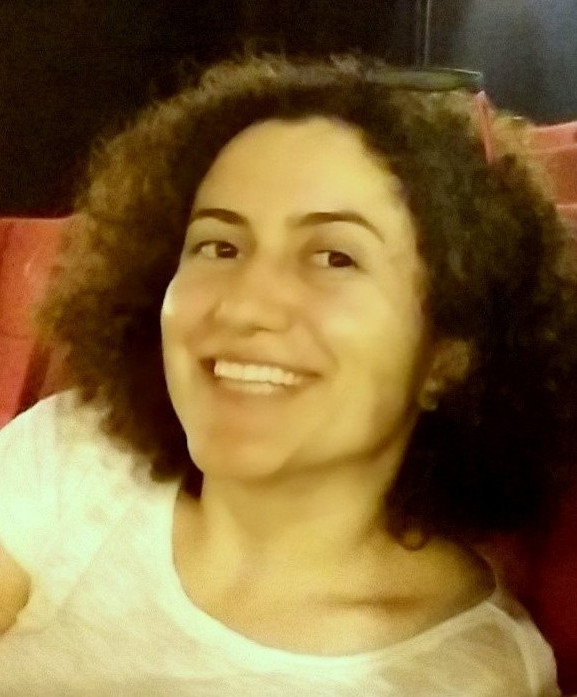}}]{Fadime Sener}
received her BSc.\ in Computer Engineering in 2011 from Hacettepe University and her MSc.\ degree in Computer Engineering from Bilkent University in 2013. In 2015, she joined the Visual Computing Group of Angela Yao at the Institute of Computer Science at the University of Bonn as a doctoral researcher and received her PhD in 2021. Her research focuses on automatic human activity understanding in videos through vision and language.
\end{IEEEbiography}

\begin{IEEEbiography}[{\includegraphics[width=1in,height=1.25in,clip,
keepaspectratio]{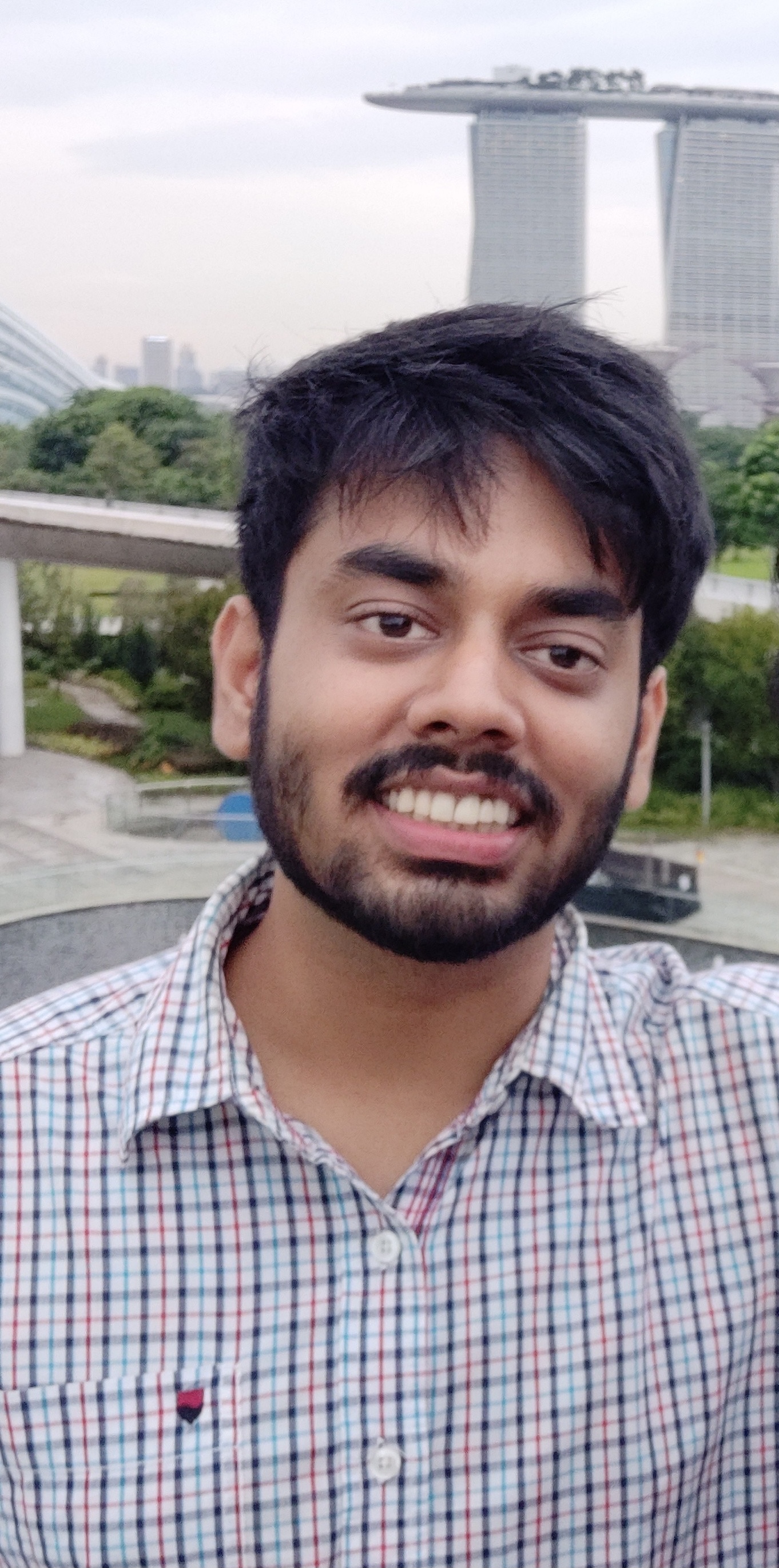}}]{Rishabh Saraf}
received his Integrated M.Tech. in Mathematics and Computing from the Indian Institute of Technology Dhanbad. He is currently working as a data scientist at Rakuten, Inc. In 2020, he was an intern at IBM India Software Labs. In 2019, he was an intern in the CVML group of Angela Yao at NUS. In 2018, he was an intern at the University of Manitoba. He worked on various projects in computer vision, NLP, mathematical statistics, entity matching and linking for master data management.
\end{IEEEbiography}
\vspace{-9mm}

\begin{IEEEbiography}[{\includegraphics[width=1in,height=1.25in,clip,
keepaspectratio]{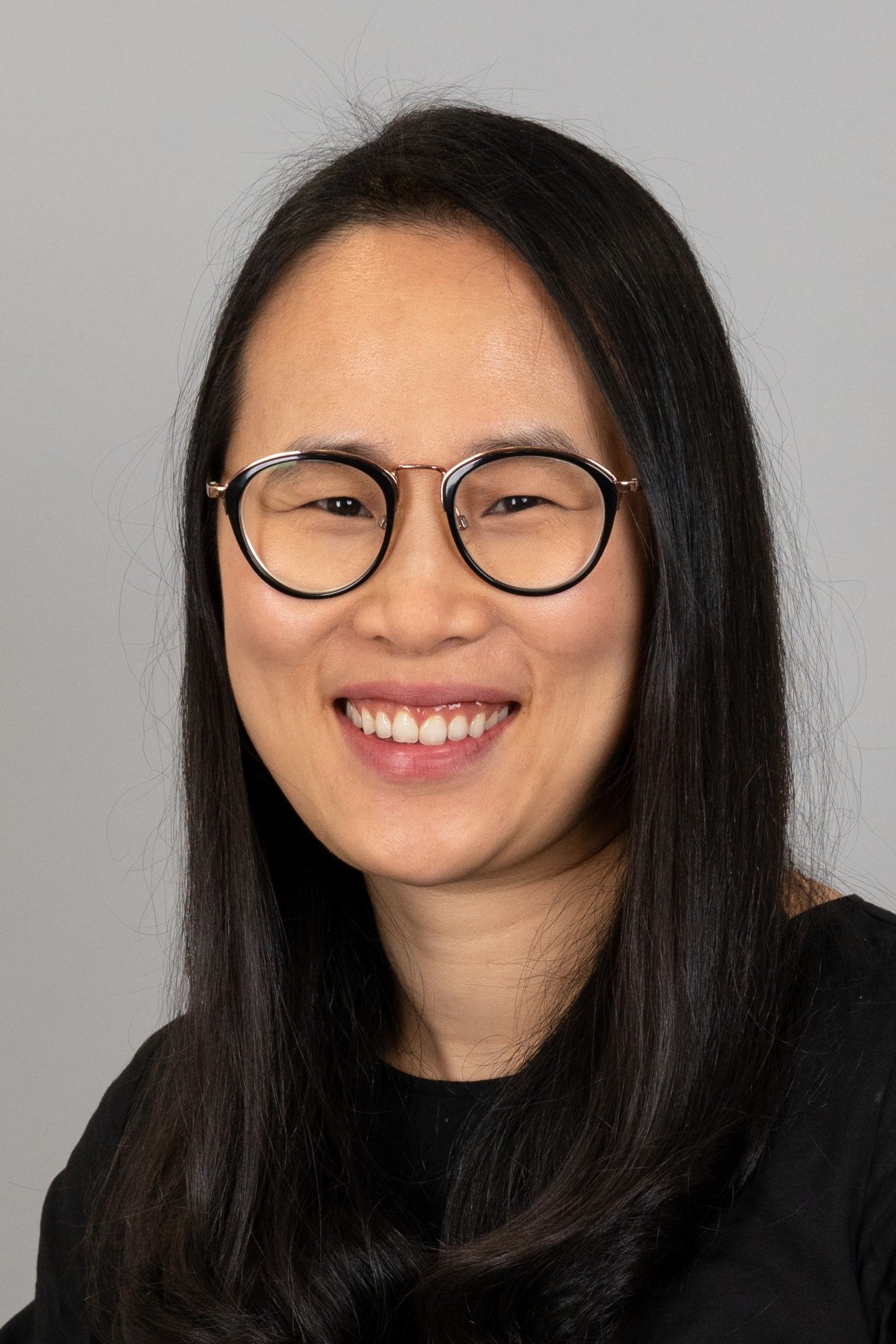}}]{Angela Yao}
received her B.A.Sc. in Engineering Science from the University of Toronto in 2006 and Master’s and PhD in 2008 and 2012 respectively from ETH Zurich. Since 2018, she is an assistant professor in the School of Computing at the National University of Singapore, where she leads the Computer Vision and Machine Learning group. Her group’s research ranges from low-level enhancement to high-level semantic interpretation of images and video. 
\end{IEEEbiography}

\end{document}